\documentclass{article}

\usepackage[final]{neurips_2024}

\usepackage[utf8]{inputenc} 
\usepackage[T1]{fontenc}    
\usepackage[colorlinks=true, linkcolor=blue, urlcolor=blue, citecolor=blue]{hyperref}
\usepackage{url}            
\usepackage{booktabs}       
\usepackage{amsfonts}       
\usepackage{nicefrac}       
\usepackage{microtype}      
\usepackage{xcolor}         
\usepackage{graphicx}
\usepackage{amsmath}
\usepackage{amsthm}
\usepackage[noend]{algpseudocode}
\usepackage{algorithm}
\usepackage{tikz}

\newcommand{\xRightarrow}[2][]{\ext@arrow 0359\Rightarrowfill@{#1}{#2}}
\graphicspath{{./figures/}}

\newtheorem{lemma}{Lemma}
\theoremstyle{definition}
\newtheorem{hypothesis}{Hypothesis}
\title{RRAEDy: Adaptive Latent Linearization of Nonlinear Dynamical Systems}

\author{%
  Jad Mounayer\\
  ENSAM Institute of Technology\\
  PIMM, SKF Chair\\
  Blvd de l'Hôpital, Paris \\
  \texttt{jad.mounayer@ensam.eu}\\
  \And
  Sebastian Rodriguez\\
  ENSAM Institute of Technology\\
  PIMM, RTE Chair\\
  Blvd de l'Hôpital, Paris \\
  \texttt{sebastian.rodriguez\_iturra@ensam.eu}\\
  \And
  Jerome Tomezyk\\
  SKF Magnetic Mechatronics\\
  Rue des Champs, Saint-Marcel\\
  \texttt{jerome.tomezyk@skf.com}\\
  \And
  Chady Ghnatios \\
  University of North Florida\\
  1 UNF Dr., Jacksonville\\
  \texttt{chady.ghnatios@unf.edu}\\
  \And
  Francisco Chinesta \\
  ENSAM Institute of Technology\\
  CNRS@CREATE, Singapore\\
  Blvd de l'Hôpital, Paris\\
  \texttt{francisco.chinesta@ensam.eu} \\
}

\begin{document}

\maketitle

\begin{abstract}
  Most existing latent-space models for dynamical systems require fixing the latent dimension in advance, they rely on complex loss balancing to approximate linear dynamics, and they don't regularize the latent variables. We introduce RRAEDy, a model that removes these limitations by discovering the appropriate latent dimension, while enforcing both regularized and linearized dynamics in the latent space. Built upon Rank-Reduction Autoencoders (RRAEs), RRAEDy automatically rank and prune latent variables through their singular values while learning a latent Dynamic Mode Decomposition (DMD) operator that governs their temporal progression. This structure-free yet linearly constrained formulation enables the model to learn stable and low-dimensional dynamics without auxiliary losses or manual tuning. We provide theoretical analysis demonstrating the stability of the learned operator and showcase the generality of our model by proposing an extension that handles parametric ODEs. Experiments on canonical benchmarks, including the Van der Pol oscillator, Burgers’ equation, 2D Navier–Stokes, and Rotating Gaussians, show that RRAEDy achieve accurate and robust predictions. Our code is open-source and available \href{https://github.com/JadM133/RRAEDy}{here}. We also provide a video summarizing the main results \href{https://youtu.be/ox70mSSMGrM}{here}.
\end{abstract}


\section{Introduction}

Dynamical systems underlie a wide range of natural and engineered processes, like the cooling of a cup of coffee or the motion of a vehicle. Many such systems can be described (or approximated) by an ordinary differential equation (ODE)
\begin{equation*}\label{eq:ode}
    \dot{\mathbf{x}} = f(\mathbf{x}, t), 
    \qquad \text{or equivalently} \qquad 
    \mathbf{x}_{t+1} = \tilde f(\mathbf{x}_t, t),
\end{equation*}
where $\mathbf{x}$ denotes the system’s state and \(f\) its governing dynamics. When \(f\) is known, numerical solvers can predict future states or simulate new trajectories. In practice, however, \(f\) is rarely known and must be inferred from data.

\paragraph{Classical system identification.}
Classical methods estimate \(f\) under structural assumptions. Dynamic Mode Decomposition (DMD)~\cite{tu2013dmd} assumes linearity, while Sparse Identification of Nonlinear Dynamical Systems (SINDy)~\cite{brunton2016sindy} seeks a sparse combination of predefined basis functions. Symbolic regression and autoregressive formulations provide further flexibility, but all depend on strong prior assumptions about the functional form of \(f\).

\paragraph{Neural approaches.}
To overcome these constraints, neural approaches replace \(f\) with a neural network, exploiting its universal approximation ability. Continuous-time variants include Neural ODEs~\cite{chen2019neuralordinarydifferentialequations} and Neural Controlled Differential Equations (Neural CDEs)~\cite{kidger2020ncde}, which learn smooth latent trajectories by integrating a neural network over time. Discrete-time counterparts include Recurrent Neural Networks (RNNs), Long Short-Term Memory networks (LSTMs), and Gated Recurrent Units (GRUs)~\cite{chung2014empiricalevaluationgatedrecurrent, lstm, schmidt2019rnn}, as well as Temporal Convolutional Networks (TCNs)~\cite{lea2016temporalconvolutionalnetworksunified}. Other models have been used for different applications related to time series such as Graph Neural Networks (GNNs)~\cite{tierz2025gnn, zhou2021graphneuralnetworksreview}, or Physics-Informed Neural Networks (PINNs)~\cite{park2025pintphysicsinformedneuraltime, raissi2017physicsinformeddeeplearning}. In addition, some models have been developed specifically for learning dynamical systems, such as Hamiltonian Neural Networks (HNNs)~\cite{greydanus2019hnn}, or Lagrangian Neural Networks (LNNs)~\cite{cranmer2020lagrangianneuralnetworks}.

While these neural formulations are highly expressive, they often exhibit practical difficulties. In continuous-time settings, gradients may vanish or explode~\cite{fronk2025vanishinggradientproblemstiff}, and optimization can converge to poor local minima \cite{ghazal2025applicationreducedordermodelstemporal, thummerer2023eigeninformedneuralodesdealingstability}. Moreover, without explicit structure or constraints, purely neural models tend to overfit and yield unstable long-term forecasts \cite{gal2016theoreticallygroundedapplicationdropout}.

\paragraph{Latent-space modeling.}
Because learning directly in the high-dimensional data space can be inefficient, many works learn dynamics in a \emph{latent space} where evolution is simpler. Latent ODEs~\cite{rubanova2019latentodesirregularlysampledtime} use a variational autoencoder to define this space, while Recurrent Autoencoder Networks (RAEs)~\cite{Susik_2021} couple autoencoders with recurrent cells. Despite good empirical results, these models usually require manual tuning of the latent dimension, and do not resolve the overfitting issue since no structure is imposed on the dynamics.

\paragraph{Koopman-inspired models.}
From Koopman theory~\cite{arbabi2016koopman, brunton2021modernkoopmantheorydynamical}, any nonlinear dynamical system can be represented linearly in a suitably chosen function space. This idea inspired methods such as the Deep Learning Enhanced DMD~\cite{Alford_Lago_2022}, DMD Autoencoders~\cite{saha2024bridgingautoencodersdynamicmode}, and Koopman Autoencoders~\cite{Lusch_2018}, which learn an autoencoded latent representation with linear time evolution. However, these approaches (i) fix the latent dimension a priori, which is problematic when the system’s intrinsic dimensionality is unknown, (ii) do not regularize their latent spaces, so they might struggle on data with local behavior, and (iii) rely on delicate weighting of multiple losses, which hampers training stability.

\paragraph{Our approach.}
To address these challenges, we build upon Rank Reduction Autoencoders (RRAEs)~\cite{mounayer2025rankreductionautoencoders}, which adaptively determine the latent dimension via a truncated Singular Value Decomposition (SVD) in the latent space. RRAEs enforce a low-rank structure without additional loss terms, improving stability and interpretability. Both RRAEs, and their variants \cite{ResearchSquare2025, mounayer2025variationalrankreductionautoencoders} show that some latent properties of Autoencoders can be imposed without complicating the loss function.

Building on this principle, we introduce \textbf{RRAEDy}, an autoencoder architecture that learns the appropriate latent dimension, while enforcing regularization and linearization on the latent dynamics. RRAEDy combine the adaptive bottleneck of aRRAEs with a dynamically learned DMD operator that governs latent evolution. By construction, this enforces linear dynamics without auxiliary loss balancing and improves stability during training and prediction.

\paragraph{We summarize our contributions as follows:}
\begin{itemize}
  \item \textbf{Methodology:} We propose RRAEDy, a model that jointly adapts the latent dimension and enforces regularized dynamics that can be linearized through a learned DMD operator, without additional loss terms.
  \item \textbf{Architecture:} We design an end-to-end architecture combining an encoder, decoder, latent SVD, and latent DMD, enabling efficient learning of dynamical systems from time series data. The architecture of both the encoder and the decoder does not depend on the dimension of the latent dynamics. Optimization is performed by minimizing a single reconstruction loss.
  \item \textbf{Theory:} We provide stability analysis and show that the learned operator converges to the identity operator, promoting smooth training dynamics. We also show that the learned DMD operartors for different batches converge to similar operators.
  \item \textbf{Empirics:} We validate RRAEDy on benchmark systems (Van der Pol oscillator, Burgers’ equation, 2D Navier--Stokes, and Rotation of Gaussians), demonstrating accurate forecasting, stable behavior, and robustness to local behavior.
  \item \textbf{Ablation Study:} We perform a detailed ablation study where we validate the theory provided in the methodology section, and we show the importance of each component of our model.
  \item \textbf{Parametric ODEs:} We extend RRAEDy to handle parametric ODEs, demonstrating the model's flexibility in learning dynamics across varying system parameters.
  \item \textbf{Reproducibility:} We provide an open-source implementation of RRAEDy, along with scripts to reproduce all experiments in Equinox \cite{kidger2021equinox}, JAX \cite{jax2018github}, available \href{https://github.com/JadM133/RRAEDy}{here}. We also created a video summarizing the main results, available \href{https://youtu.be/ox70mSSMGrM}{here}.
\end{itemize}

\section{Background}
Our model is based on both Rank Reduction Autoencoders (RRAEs) \cite{mounayer2025rankreductionautoencoders} and the Dynamic Mode Decomposition (DMD) \cite{tu2013dmd}. In this section, we present both of these with our own notations.

\underline{\textbf{Rank Reduction Autoencoders:}}

Since RRAEs are based on the Singular Value Decomposition (SVD), we start by defining the SVD. Any matrix $Y \in \mathbb{R}^{m \times n}$ can be decomposed as follows,
\begin{equation*}
    Y = U\Sigma V^T = U\alpha = \sum_{i=1}^{n}U_i\sigma_iV_i^T,
\end{equation*}
where $U\in\mathbb{R}^{m \times n}$, $\Sigma\in\mathbb{R}^{n \times n}$, and $V \in \mathbb{R}^{n \times n}$. An approximation of $Y$ can be obtained by truncating the sum to the first $k^* \leq n$ terms, which is known as the truncated SVD,
\begin{equation*}
    Y^{(k^*)} = U^{(k^*)}\Sigma^{(k^*)}\left(V^{(k^*)}\right)^T =  U^{(k^*)}\alpha^{(k^*)} = \sum_{i=1}^{k^*}U_{:, i}\sigma_iV_{:, i}^T,
\end{equation*}
where $U^{(k^*)} \in\mathbb{R}^{L\times k^*}$ and $V^{(k^*)} \in\mathbb{R}^{n\times k^*}$ contain the first $k^*$ columns of $U$ and $V$ respectively, $\Sigma^{(k^*)}$ is a diagonal matrix containing the first $k^*$ sorted singular values $[\sigma_1 \geq \dots \geq \sigma_{k^*}]$, and both $U_{:, i}$ and $V_{:, i}$ are columns of $U$ and $V$ respectively. Note that $\alpha^{(k^*)} = \Sigma^{(k^*)}\left(V^{(k^*)}\right)^T \in\mathbb{R}^{k^*\times n}$ represents a compressed representation of $Y$ if the basis is known.

Let $X = [x_1, \dots, x_N]\in\mathbb{R}^{F\times N}$, where each $x_i\in\mathbb{R}^{F}$, let $k^*$ be a chosen bottleneck size, and $L >k^*$ be a chosen latent size. An RRAE can be trained as follows,
\begin{equation*}
  \begin{cases}
    \text{\underline{Encoding:}}\\[2ex]
    Y_{:, j}= \mathcal{E}(X_{:, j}) \quad \forall j\in[1, N], \quad &\text{with,} \quad Y \in \mathbb{R}^{L\times N}, \qquad \mathcal{E}: \mathbb{R}^{F} \xrightarrow{} \mathbb{R}^{L},\\[2ex]
    \text{\underline{Latent SVD:}}\\[2ex]
    Y^{(k^*)} = \displaystyle\underbrace{\sum_{i=1}^{k^*}U_{:, i}\sigma_iV_{:, i}^T}_{\text{Truncated SVD}} = U^{(k^*)}\alpha^{(k^*)}, \quad &\text{with,} \quad U^{(k^*)} \in\mathbb{R}^{L\times k^*}, \qquad \alpha^{(k^*)} \in\mathbb{R}^{k^* \times N}, \\[8ex]
    \text{\underline{Decoding:}}\\[2ex]
    \tilde{X}_{:, j}= \mathcal{D}\left(Y^{(k^*)}_{:, j}\right) \quad \forall j\in[1, N], \quad &\text{with,} \quad \tilde{X} \in \mathbb{R}^{F\times N}, \qquad \mathcal{D}: \mathbb{R}^{L} \xrightarrow{} \mathbb{R}^{F}.\\[2ex]
  \end{cases}
\end{equation*}

Where both $\mathcal{E}$ and $\mathcal{D}$ are neural networks. We use the notation $M_{:, i}$ to denote the $i^{th}$ column of a matrix $M$. RRAEs only minimize the reconstruction loss $\|\tilde{X}-X\|_*$ (any norm), and they regularize their latent space by construction using the SVD.

Further, note that during inference, the SVD is replaced by a projection onto a common basis $U_f$ found by a final SVD on the bases of all batches, as proposed by \cite{mounayer2025rankreductionautoencoders}. Accordingly, during inference, the model can encode and decode any new sample $X_l$ as follows,

\begin{equation*}
  \begin{cases}
    \text{\underline{Encoding:}}\\[2ex]
    Y_l = \mathcal{E}(X_l), \,  &\text{with,} \quad Y_l \in \mathbb{R}^{L\times 1}, \qquad \mathcal{E}: \mathbb{R}^{F} \xrightarrow{} \mathbb{R}^{L},\\[2ex]
    \text{\underline{Latent SVD (projection):}}\\[2ex]
    \alpha_l^{(k^*)} =  U_f^TY_l, \quad &\text{with,} \quad U_f \in\mathbb{R}^{L\times k^*}, \quad \alpha^{(k^*)}_l \in\mathbb{R}^{k^* \times 1}, \\[2ex]
    Y^{(k^*)}_l = U_f\alpha^{(k^*)}_l, \quad &\text{with,} \quad Y^{(k^*)}_l \in\mathbb{R}^{L\times 1}, \\[2ex]
    \text{\underline{Decoding:}}\\[2ex]
    \tilde{X}_{l}= \mathcal{D}\left(Y^{(k^*)}_l\right), \quad &\text{with,} \quad \tilde{X}_l \in \mathbb{R}^{F\times 1}, \qquad \mathcal{D}: \mathbb{R}^{L} \xrightarrow{} \mathbb{R}^{F}.\\[1ex]
  \end{cases}
\end{equation*}

The main advantage of using RRAEs is that the bottleneck $\alpha^{(k^*)}$ has a specific structure (sorted diagonal multiplying a truncated orthonormal matrix). In later sections, we explain how we can leverage this structure to ensure the stability of the learned dynamics.

\underline{\textbf{Dynamic Mode Decomposition:}}

The Dynamic Mode Decomposition (DMD) is a data-driven technique used to learn a linear operator that best fits the evolution of a dynamical system. Let $Z = [z_1, \dots, z_T] \in\mathbb{R}^{P\times T}$ be a sequence of observations of a dynamical system at different times, where each $z_i\in\mathbb{R}^{P}$. DMD aims to find the best linear operator $W\in\mathbb{R}^{P\times P}$ such that,
\begin{equation*}
    z_{t+1} \approx Wz_t, \quad \forall t\in[1, T-1].
\end{equation*}
By defining $Z_p = Z_{:, 2:T} = [z_2, \dots, z_T]$ and $Z_m = Z_{:, 1:T-1} =[z_1, \dots, z_{T-1}]$, the DMD operator can be found as follows,
\begin{equation*}
    W = \arg\min_{A} ||Z_p - AZ_m||_F^2 = Z_p\left(Z_m\right)^{\dagger},
\end{equation*}
where $||.||_F$ is the Frobenius norm and $^\dagger$ is the Moore-Penrose pseudoinverse. Finding the DMD operator is closely related to the SVD. By computing the SVD of $Z_m$, $Z_m = U_{Z_m}\Sigma_{Z_m} V^T_{Z_m}$, the DMD operator can be expressed as follows,
\begin{equation*}
    W = Z_pV_{Z_m}\Sigma^{-1}_{Z_m}U^T_{Z_m}.
\end{equation*}
Using the operator $W$, one can find a prediction at all times as follows,
\begin{equation*}
    Z_{DMD} \approx Z = \left[z_1, Wz_1, W^2z_1, \dots, W^{T-1}z_1\right].
\end{equation*}

\section{Methodology}\label{sec:metho}
In this section, we present our model, Rank Reduction Autoencoders for Dynamical Systems (RRAEDy). We start by defining the model, then we present some of its theoretical and practical advantages.

\begin{figure}[!h]
    \centering
    \includegraphics[width=1\textwidth, trim=0 4.5cm 0 12.5cm, clip]{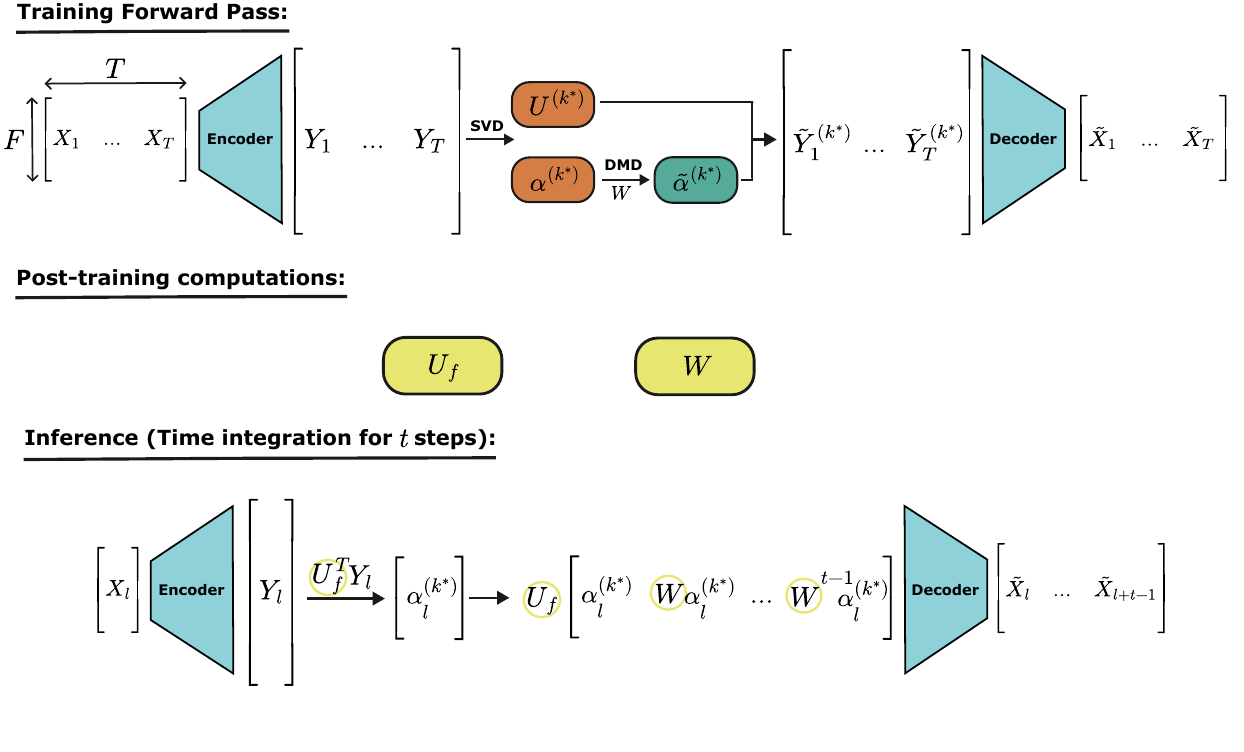}
    \caption{Architecture of RRAEDy for one sample ($N=1$).}
    \label{fig:dyn_arrae_diagram}
\end{figure}

The model consists of combining both RRAEs and the DMD to learn time series using Autoencoders. Without loss of generality\footnote{For other types of data such as images, the model is defined similarly and the encoder/decoder would include flatten/unflatten operations.}, let $X\in\mathbb{R}^{F\times T}$ denote the data matrix containing one time series of length $T$, where each observation is a vector of size $F$. Throughout this section and for simplicity, we assume that we only have one time series (i.e., $N=1$, one sample).  In practice, when $N$ samples are available, the input becomes a tensor of shape $(F \times T \times N)$. In this case, the model is slightly modified by including some flattening/unflattening operations, as described in Appendix \ref{ap:architecture}.

\paragraph{Model definition for a fixed bottleneck.} For a chosen bottleneck size $k^*$, and a chosen latent size $L >k^*$, as shown in Figure \ref{fig:dyn_arrae_diagram}, we define RRAEDy during training as follows,

\begin{equation*}
  \begin{cases}
    \text{\underline{Encoding:}}\\[2ex]
    Y_{:, j}= \mathcal{E}(X_{:, j}), \quad \forall j\in[1, T], \,  &\text{with,} \quad Y \in \mathbb{R}^{L\times T}, \qquad \mathcal{E}: \mathbb{R}^{F} \xrightarrow{} \mathbb{R}^{L},\\[2ex]
    \text{\underline{Latent SVD:}}\\[2ex]
    Y^{(k^*)} = \displaystyle\underbrace{\sum_{i=1}^{k^*}U_{:, i}\sigma_iV_{:, i}^T}_{\text{Truncated SVD}} = U^{(k^*)}\alpha^{(k^*)}, \quad &\text{with,} \quad U^{(k^*)} \in\mathbb{R}^{L\times k^*}, \quad \alpha^{(k^*)} \in\mathbb{R}^{k^* \times T}, \\[8ex]
    \text{\underline{Latent DMD:}}\\[2ex]
    \alpha^{(k^*)}_m = \alpha_{:, 1:T-1}^{(k^*)}, \qquad \alpha^{(k^*)}_p = \alpha_{:, 2:T}^{(k^*)}, \quad &\text{with,} \quad \alpha^{(k^*)}_m, \alpha^{(k^*)}_p \in\mathbb{R}^{k^*\times T-1},  \\[2ex]
    W = \alpha^{(k^*)}_p \left(\alpha^{(k^*)}_m\right)^{\dagger}, & \text{with,}\quad W\in\mathbb{R}^{k^*\times k^*},\\[2ex]
    \alpha^{(k^*)}_{DMD} = \left[\alpha^{(k^*)}_{:, 1}, W\alpha^{(k^*)}_{:, 1}, \dots, W^{T-1}\alpha^{(k^*)}_{:, 1}\right],\\[2ex]
    \text{\underline{Latent Reconstruction:}}\\[2ex]
    \tilde{Y}^{(k^*)} = U^{(k^*)}\alpha^{(k^*)}_{DMD}, \quad &\text{with,} \quad \tilde{Y}^{(k^*)} \in\mathbb{R}^{L\times T}, \\[2ex]
    \text{\underline{Decoding:}}\\[2ex]
    \tilde{X}_{:, i}= \mathcal{D}\left(\tilde{Y}^{(k^*)}_{:, i}\right) \quad \forall i\in[1, T], \quad &\text{with,} \quad \tilde{X} \in \mathbb{R}^{F\times T}, \qquad \mathcal{D}: \mathbb{R}^{L} \xrightarrow{} \mathbb{R}^{F},\\[1ex]
  \end{cases}
\end{equation*}
where again we use the notation $M_{:, i}$ to denote the $i^{th}$ column of a matrix $M$. The model, when trained, consists of five main steps: (i) encoding the data into a latent space using an encoder $\mathcal{E}$, (ii) applying a truncated SVD to the latent representation to obtain a compressed representation $\alpha^{(k^*)}$ (RRAE bottleneck), (iii) learning a DMD operator $W$ that governs the linear evolution of $\alpha^{(k^*)}$ over time, (iv) reconstructing the latent representation using the learned DMD operator (initial value problem using the DMD), and (v) decoding the reconstructed latent representation back to the original space using a decoder $\mathcal{D}$.

On the other hand, both the DMD and the SVD in the latent space have different behaviors during inference. Once training is done, a common basis $U_f$ (found as proposed by \cite{mounayer2025rankreductionautoencoders}, and as discussed previously in Section \ref{sec:metho}) is fixed and the SVD is replaced by a projection onto this basis. The DMD operator $W$ is computed over the whole training dataset, and also fixed and used to predict the evolution of the latent coefficients. Accordingly, during inference, the model can evolve the system from any point in time $X_l$ for $t$ timesteps as follows,

\begin{equation*}
  \begin{cases}
    \text{\underline{Encoding:}}\\[2ex]
    Y_l = \mathcal{E}(X_l), \,  &\text{with,} \quad Y_l \in \mathbb{R}^{L\times 1}, \qquad \mathcal{E}: \mathbb{R}^{F} \xrightarrow{} \mathbb{R}^{L},\\[2ex]
    \text{\underline{Latent SVD (projection):}}\\[2ex]
    \alpha_l^{(k^*)} =  U_f^TY_l, \quad &\text{with,} \quad U_f \in\mathbb{R}^{L\times k^*}, \quad \alpha^{(k^*)}_l \in\mathbb{R}^{k^* \times 1}, \\[2ex]
    \text{\underline{Latent DMD (IVP using $W$):}}\\[2ex]
    \alpha^{(k^*)}_{DMD} = \left[\alpha^{(k^*)}_l, W\alpha^{(k^*)}_l, \dots, W^{t-1}\alpha^{(k^*)}_l\right], &\text{with,} \quad \alpha^{(k^*)}_{DMD} \in \mathbb{R}^{k^*\times t},\\[2ex]
    \text{\underline{Latent Reconstruction:}}\\[2ex]
    \tilde{Y}^{(k^*)} = U_f\alpha^{(k^*)}_{DMD}, \quad &\text{with,} \quad \tilde{Y}^{(k^*)} \in\mathbb{R}^{L\times t}, \\[2ex]
    \text{\underline{Decoding:}}\\[2ex]
    \tilde{X}_{:, i}= \mathcal{D}\left(\tilde{Y}^{(k^*)}_{:, i}\right) \quad \forall i\in[1, t], \quad &\text{with,} \quad \tilde{X} \in \mathbb{R}^{F\times t}, \qquad \mathcal{D}: \mathbb{R}^{L} \xrightarrow{} \mathbb{R}^{F},\\[1ex]
  \end{cases}
\end{equation*}

\paragraph{Advantages (Theoretical and Practical).} RRAEDy present several advantages compared to other models used for learning dynamical systems from time series. We summarize these advantages below.

\underline{\textbf{The dimension of the latent dynamics: Adaptive selection.}} Note that the bottleneck $\alpha^{(k^*)}$, which represents our latent dynamical system, can be written as $\alpha^{(k^*)} = \Sigma^{(k^*)}\left(V^{(k^*)}\right)^T$, with $\Sigma^{(k^*)}$ a diagonal matrix with sorted values, and $\left(V^{(k^*)}\right)^T$ an orthonormal matrix. Accordingly, the significance of the latent variables is known a priori. Further, a main advantage of RRAEDy is that the architecture of both the encoder and the decoder does not depend on the bottleneck size $k^*$. Both of these properties allows us to adaptively select $k^*$ during training, as proposed in \cite{mounayer2025rankreductionautoencoders}. The algorithm can be found below,

\begin{algorithm}[H]
  \caption{Adaptive Rank Reduction Autoencoder Algorithm (modified from \cite{mounayer2025rankreductionautoencoders})}
  \label{alg:adaptive}
  \begin{algorithmic}
    \State \textbf{Input:} Unlabeled Data $X \in \mathbb{R}^{F\times T}$, stagnation criteria.
    \State Initialize $k^*= \min (T, L) $.
    \State Train RRAEs until convergence.
    \State Store the final epoch's loss as the \emph{optimal\_loss}.
    \While {True}
      \State $k^*=k^*-1$.
      \State Continue training the RRAE with the new $k^*$ until one of the following,
      
      \If {\emph{optimal\_loss} reached}
        \State Save model at $k^*$.
      \ElsIf {stagnated}
        \State $k^*=k^*+1$.
        \State Load model saved at $k^*$.
        \State Break.

      \EndIf
    \EndWhile

    \State Find the final training basis $U_f$.
    \State Find the final DMD operator $W$ over the whole dataset.
    \State \textbf{Output:} Trained RRAE with the optimal bottleneck size, $U_f \in \mathbb{R}^{L\times k^*}$, and $W \in \mathbb{R}^{k^*\times k^*}$.
  \end{algorithmic}
\end{algorithm}

The presented algorithm eliminates the need to choose $k^*$ a priori, which is often challenging when the system's intrinsic dimensionality is unknown.

\underline{\textbf{Regularization: Robustness to local behavior.}} The latent dynamics in RRAEDy are regularized by construction due to the SVD structure of the bottleneck with $\alpha^{(k^*)} = \Sigma^{(k^*)}\left(V^{(k^*)}\right)^T$, and $V^{(k^*)}$ being an orthonormal matrix (i.e., the sum of the squares of its elements in each column is $1$). This regularization promotes robustness to datasets with local behavior as demonstrated in both \cite{mounayer2025rankreductionautoencoders,mounayer2025variationalrankreductionautoencoders}. This is a significant advantage over other models that do not impose any structure on the latent dynamics, leading to what is known as ``holes'' in the latent space, illustrated by overfitting and poor performance on the test set, as shown in some samples in Section \ref{sec:ablation}.

\underline{\textbf{The objective function: One loss term.}} The model is trained by minimizing the reconstruction loss between the input $X$ and the output $\tilde{X}$, using a single loss term:
\begin{equation*}
    \mathcal{L}(X, \tilde{X}) = \|X - \tilde{X}\|_*,
\end{equation*}
where $\|\cdot\|_*$ is any norm of our choice. Accordingly, RRAEDy are training an encoder, a decoder, finding the latent dimension, and enforcing linear dynamics in the latent space, all by minimizing a single loss term. This is a significant advantage over other models that require balancing multiple loss terms, which can complicate optimization and lead to suboptimal convergence. Since both the DMD and the SVD provide approximations in the least-square sense, the loss term above enforces the latent coefficients $\alpha^{(k^*)}$ to be well approximated by linear dynamics governed by the DMD operator $W$.

\underline{\textbf{Theoretical Guarantees: Similarity and Marginal stability.}} We provide two theoretical guarantees regarding the learned DMD operator $W$. First, we show that when training using batches, the learned DMD operators for different batches converge to similar operators. Second, we show that the learned operator remains close to identity during training, which promotes smooth training dynamics. We start by defining three hypotheses that we will use in the following analysis.

\begin{hypothesis}\label{hyp:1}
    The number of timesteps $T$ is large enough ($T \gg k^*$) to ensure that we can capture the dynamics of the system. The entries are also i.i.d, and the dynamics are varied enough in time (at least $k^*$ linearly independent states in the latent space).
\end{hypothesis}

This hypothesis is not too restrictive in practice, since time series data usually contain many timesteps and the bottleneck $k^*$ is often much smaller than $T$.

\begin{hypothesis}\label{hyp:2}
    The Autoencoder has converged, meaning that the reconstruction $\tilde{X}$ is close to the input $X$. Accordingly, both encoder $\mathcal{E}$ and decoder $\mathcal{D}$ do not change significantly with further training between epochs.
\end{hypothesis}

This hypothesis is reasonable since we are interested in the behavior of the learned DMD operator at convergence.

\begin{hypothesis}\label{hyp:3}
    Lipschitz continuity of the input and the encoder. In mathematical terms, $\|X_{:, t+1} - X_{:, t}\|_2 \leq L_X$, and $\|Y_{:, t+1} - Y_{:, t}\|_2 \leq L_E\|X_{:, t+1} - X_{:, t}\|_2, \forall t\in[1, T-1]$, where $L_X$ and $L_E$ are some positive constants, $X$ is the input time series, and $Y$ is the encoded time series.
\end{hypothesis}

This hypothesis is reasonable since in many physical systems, the state does not change abruptly between consecutive timesteps, especially when the sampling rate is high enough. Further, many Neural Networks that can be used as encoders are Lipschitz continuous.

\underline{\textbf{1- Similar DMD operators for different batches.}} Note that when processing multiple samples in batches, the bases in RRAEs have been proven to converge to an approximation of a linear combination of each others \cite{mounayer2025rankreductionautoencoders}. In other words, we can write,
\begin{equation}\label{eq:U_wQ}
    U_j \approx U_iQ, \qquad \text{with,}\qquad QQ^T = Q^TQ = I.
\end{equation}
Note that since the initial conditions $\alpha_{:, 1}^{(k^*)}$ are passed as they are to the decoder (i.e., not modified during the DMD), the same proof as in \cite{mounayer2025rankreductionautoencoders} can be followed so the property above still holds for RRAEDy. Based on this property, we have the following Lemma.
\begin{lemma}
    When both Hypothesis \ref{hyp:1} and Hypothesis \ref{hyp:2} hold, the learned DMD operators $W_i$ for different batches $i$ converge to similar operators. In other words, we can write,
    \begin{equation}\label{eq:Q_rel}
        W_j \approx Q^TW_iQ, \qquad \text{with,}\qquad QQ^T = Q^TQ = I.
    \end{equation}
\end{lemma}

The full proof can be found in Appendix \ref{ap:proof1}. Note that RRAEDy enforce the operators to be similar, but not the same. This is an advantage since it allows the model to adapt to different batches while still maintaining a consistent representation of the dynamics. Further, similar operators imply similar eigenvalues, hence the same stability properties for all batches. Next, we provide theoretical evidence to support the stability of the learned dynamics.

\underline{\textbf{2- Marginal Stability of the learned dynamics:}} The stability of the learned dynamics is crucial for ensuring that long-term predictions remain bounded and do not diverge. In our model, we can provide evidence to support the marginal stability of the learned DMD operator $W$ based on the structure of the bottleneck $\alpha^{(k^*)}$. We have the following Lemma,
\begin{lemma}\label{lemma:bound}
    Let Hypothesis \ref{hyp:1} and Hypothesis \ref{hyp:3} hold. Let $\kappa\left(Y^{(k^*)}\right)$, and $\sigma_{k^*}$ be the condition number, and the smallest singular value of the truncated latent space $Y^{(k^*)}$ respectively. The eigenvalues of the learned DMD operator $W$ satisfy,
    \begin{equation*}
        |\lambda_i -1| \leq \frac{\sqrt{k^*}}{\sigma_{k^*}}L_EL_X, \quad \forall i\in[1, k^*].
    \end{equation*}
    Further, we can bound the norm of $W - I$ as follows,
    \begin{equation*}
        \|W - I\|_2 \leq \frac{\sqrt{k^*}}{\sigma_{k^*}}\kappa\left(Y^{k^*}\right)L_EL_X.
    \end{equation*}

\end{lemma}

The full proof can be found in Appendix \ref{ap:proof2}. The bound above shows that the eigenvalues of the learned DMD operator $W$ remain close to $1$ during training, which promotes marginal stability.

Note that both $\sqrt{k^*}$ and $\kappa\left(Y^{(k^*)}\right)$ decrease when $k^*$ decreases. Further, $\sigma_{k^*}$ usually increases when $k^*$ decreases. According to the bound above, this means that reducing $k^*$ improves the stability of the learned dynamics. Yet, this also shows that starting training with a small $k^*$ might be restrictive since  it encourages the model to converge to $W \approx I$, which might not be sufficient to capture complex dynamics (see Section \ref{sec:ablation} for a few examples). This justifies the need to start training with a large $k^*$ and progressively reduce it, as proposed in the adaptive algorithm previously in this section.

\section{Results}
In what follows, we present numerical experiments to evaluate the performance and the convergence of RRAEDy on four time series datasets. For the readers interested in how RRAEDy extrapolate for times that are way longer than the training time span, we provide additional results in Appendix \ref{ap:extrapolation}, or in the video available \href{https://youtu.be/ox70mSSMGrM}{here}.

We begin by describing the datasets used in  our experiments.

\underline{\textbf{Van der Pol Oscillator:}} The first example is the Van der Pol oscillator, which is a non-conservative oscillator with non-linear damping. The system is described by the following second-order differential equation,
\begin{equation*}
    \frac{d^2x}{dt^2} - \mu(1 - x^2)\frac{dx}{dt} + x = 0,
\end{equation*}
where $\mu$ is a scalar parameter indicating the nonlinearity and the strength of the damping. Otherwise, the equation can be written as a system of first-order ODEs as follows,
\begin{equation*}
    \begin{cases}
        \displaystyle\frac{dx_1}{dt} = x_2,\\[3ex]
        \displaystyle\frac{dx_2}{dt} = \mu(1 - x_1^2)x_2 - x_1.
    \end{cases}
\end{equation*}
This particular system is interesting because it exhibits limit cycle behavior, where the system's trajectory converges to a closed loop in the phase space, regardless of the initial conditions. This makes it a valuable test for studying the stability and long-term behavior of dynamical systems. For our experiments, we set $\mu = 2.0$, and we generate $N=2000$ samples using random initial conditions uniformly sampled in the range $[-1.5, 1.5]$ for both $x_1$ and $x_2$. Each time series is generated over a time span of $t\in[0, 10]$ with a time step of $\Delta t = 0.05$, resulting in $T=200$ time steps per sample.

\underline{\textbf{Burger's Equation:}} The next example is the one-dimensional viscous Burger's equation, which is a fundamental partial differential equation (PDE) used to model various physical phenomena, including fluid dynamics and traffic flow. The equation is given by,
\begin{equation*}
    \frac{\partial u}{\partial t} + u\frac{\partial u}{\partial x} = \nu \frac{\partial^2 u}{\partial x^2},
\end{equation*}
where $u(x, t)$ is the velocity field, and $\nu$ is the viscosity coefficient. For our experiments, we set $\nu = 0.01$, and we generate $N=2000$ samples, discretized spacially over the domain $x\in[0, 1]$ with $F=100$ grid points, and temporally over the time span $t\in[0, 15]$ with a total of $T=400$ time steps per sample. The initial conditions for each sample are generated using polynomials of degree 4 with random coefficients, ensuring a diverse set of initial velocity profiles.

\underline{\textbf{Fluid behind a cylinder:}} The final example involves simulating the two-dimensional incompressible Navier-Stokes equations to model fluid flow behind a cylinder (similarly to the example in \cite{brunton2016sindy}). We choose this example because of its nonlinear dynamics but also to show that RRAEDy can be used with Convolutional Neural Networks (CNNs) on images as well. The equations are given by,
\begin{equation*}
    \begin{cases}
        \displaystyle\frac{\partial w}{\partial t} + (w\cdot\nabla)w = -\nabla p + \nu \nabla^2 w,\\[3ex]
        \nabla \cdot w = 0,
    \end{cases}
\end{equation*}
where $w = (u, v)$ is the velocity field, $p$ is the pressure field, and $\nu$ is the kinematic viscosity. The simulation is set up in a rectangular domain of size $2.2 \times 1$, with a circular cylinder placed at $(0.2, 0.5)$ (from the bottom corner) with a radius of $0.1$. We discretized the domain using a uniform grid with $N_x = 100$, and $N_y = 50$ grid points in the $x$ and $y$ directions, respectively, resulting in $F = N_x \times N_y = 5000$ spatial points (although, we use CNNs so the images are not flattened before being given to the network, more details about the architectures used can be found in Appendix \ref{ap:architecture}). The fluid flow is simulated over a time span of $t\in[0, 0.5]$ with a time step of $\Delta t = 0.0025$, resulting in $T=200$ time steps per sample. We generate $N=200$ samples by varying the inflow velocity profile at the left boundary between $1$ and $7$, ensuring a diverse set of flow conditions behind the cylinder.

In this case, the quantity of interest is the vorticity field $\omega$, which is computed from the velocity field as follows,
\begin{equation*}
    \omega = \frac{\partial v}{\partial x} - \frac{\partial u}{\partial y}.
\end{equation*}

\underline{\textbf{Rotation of a 2D Gaussian:}} As an additional dataset, we consider the problem of rotating a 2D Gaussian in space over time. The main purpose of using this dataset is to compare the abilities of different models on a dataset with a highly local and a cyclic behavior.

For this dataset, the initial condition is given by a Gaussian function within a 2D domain, centered on the horizontal line, and at a distance $r$ to the right from the center of the image. The Gaussian is then rotated clockwise to achive one and half circular motions around the center of the image. The governing equation for this rotation for the center of the gaussian is given by,

\begin{equation*}
    x(t) = r\cos(\omega t), \qquad y(t) = r\sin(\omega t),
\end{equation*}
where $\omega$ is the angular velocity, taking values from $0$ to $3\pi$ with $T=200$ time steps. The Gaussian function at each time step is given by,
\begin{equation*}
    G(x, y, t) = \exp\left(-\frac{(x - x(t))^2 + (y - y(t))^2}{2\sigma^2}\right),
\end{equation*}
where we set $\sigma = 0.1$. To generate the dataset, we create $N=1100$ samples by varying the radius $r$ of the initial Gaussian from $0.3$ to $0.8$ within a 2D domain with $x \in[-1,1]$ and $y \in[-1,1]$, discretized using $F=64\times64$ grid points. Finally, we normalize all the data so that the mean is $0$ and the standard deviation is $1$.

For all datasets, we split the data into training and test sets with a ratio of $80\%$ and $20\%$ respectively. We implement RRAEDy using JAX \cite{jax2018github}, specifically Equinox \cite{kidger2021equinox} and train the models using the Adabelief optimizer \cite{Zhuang2020AdaBelief}. For details about the model architectures, training hyperparameters, and additional implementation specifics, please refer to Appendix \ref{ap:architecture}.

In what follows, we present some predictions obtained using RRAEDy on the four datasets above. The input of each of the predictions presented is only the initial condition of the test time series, and the model is asked to forecast the entire time series.

\begin{figure}[h!]
    \centering
    \includegraphics[width=1\textwidth]{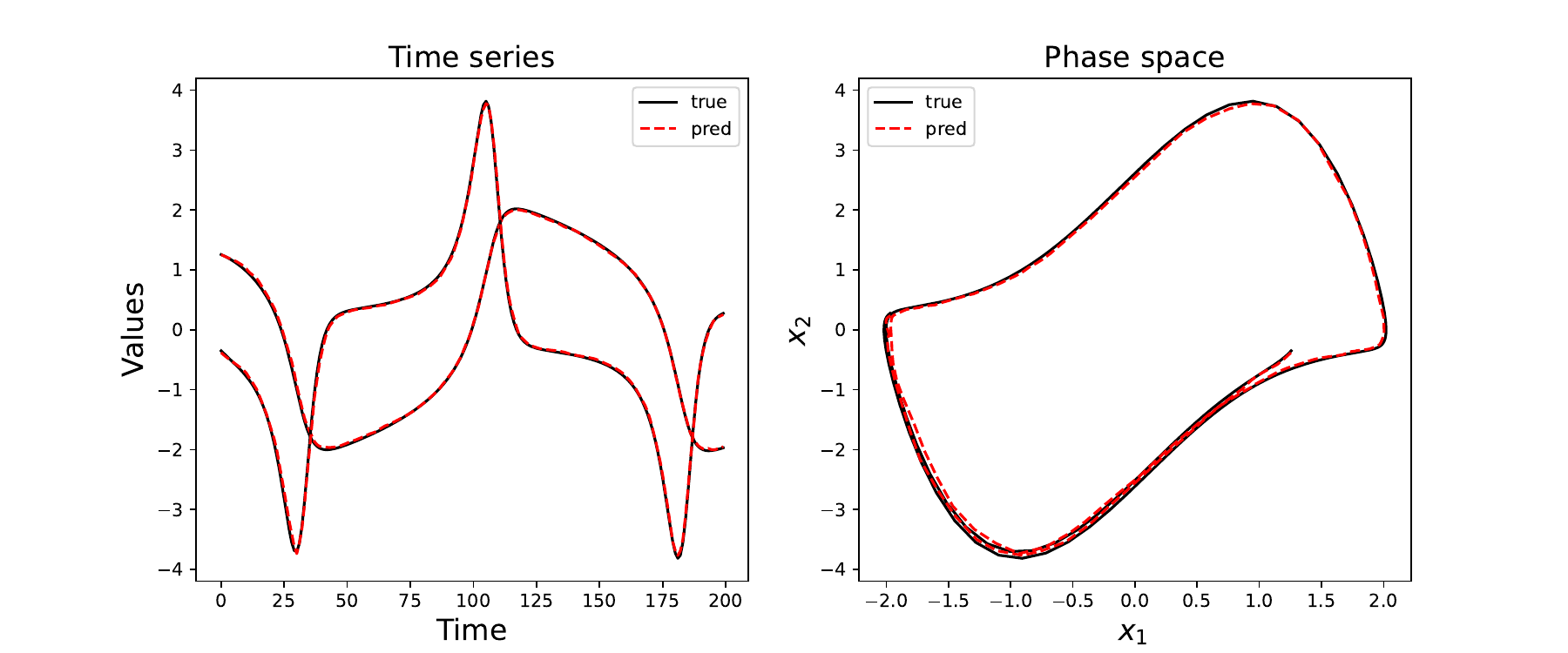}
    \caption{Predictions obtained using RRAEDy on the Van der Pol oscillator dataset (test set, only initial condition is given as input). Phase diagram (to the right), and time series (to the left).}
    \label{fig:vdp_pred}
\end{figure}

We present predictions obtained using RRAEDy on the Van der Pol oscillator in Figure \ref{fig:vdp_pred}. To the right, we show the phase space trajectory of the oscillator, while to the left, we show the time series of both states $x_1$ and $x_2$. We can see that the model is able to accurately capture the dynamics of the system.

\begin{figure}[h!]
    \centering
    \includegraphics[width=1\textwidth]{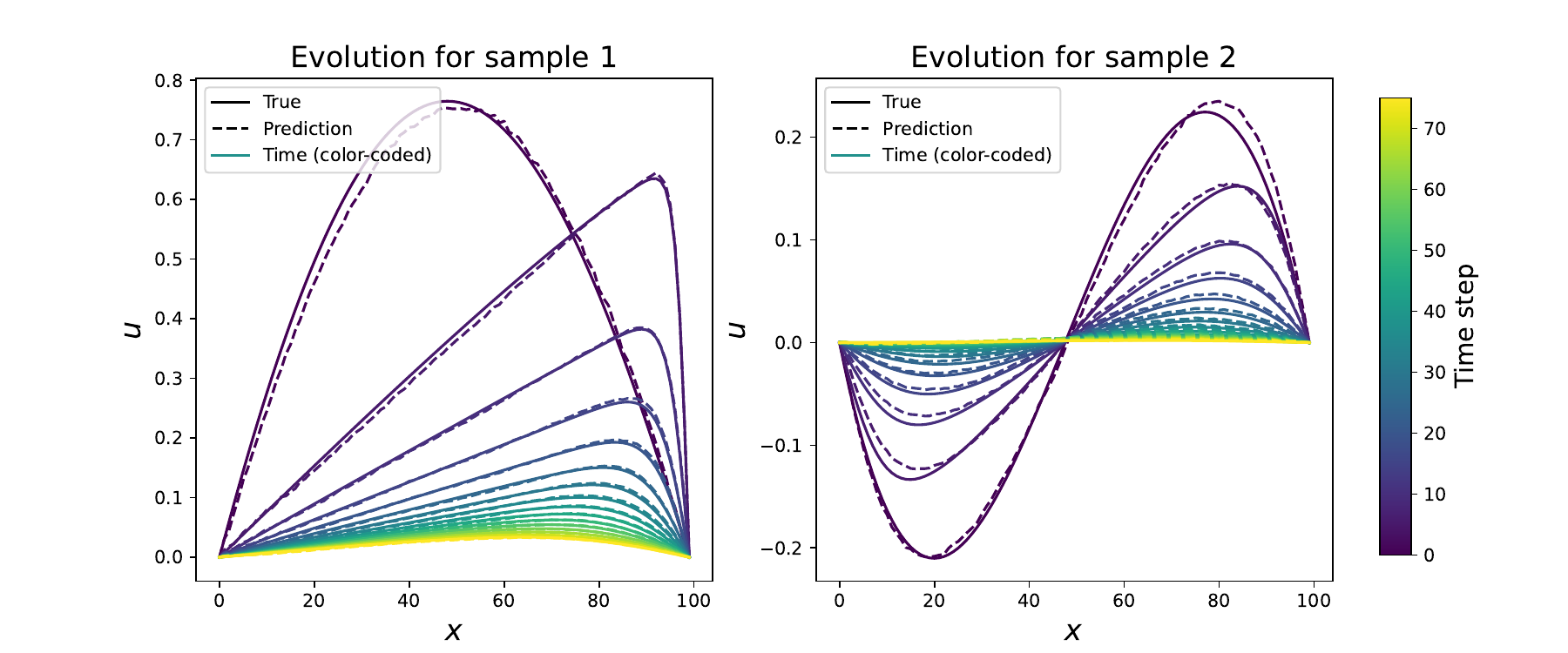}
    \caption{Predictions obtained using RRAEDy on the Burger's equation dataset for two initial conditions (test set, only initial condition is given as input).}
    \label{fig:burger_pred}
\end{figure}

Next, we present predictions obtained for two different initial conditions using RRAEDy on the Burger's equation dataset in Figure \ref{fig:burger_pred}. The model is able to accurately capture the shock formation and propagation in the solution of the Burger's equation.

\begin{figure}
    \centering
    \includegraphics[width=0.8\textwidth]{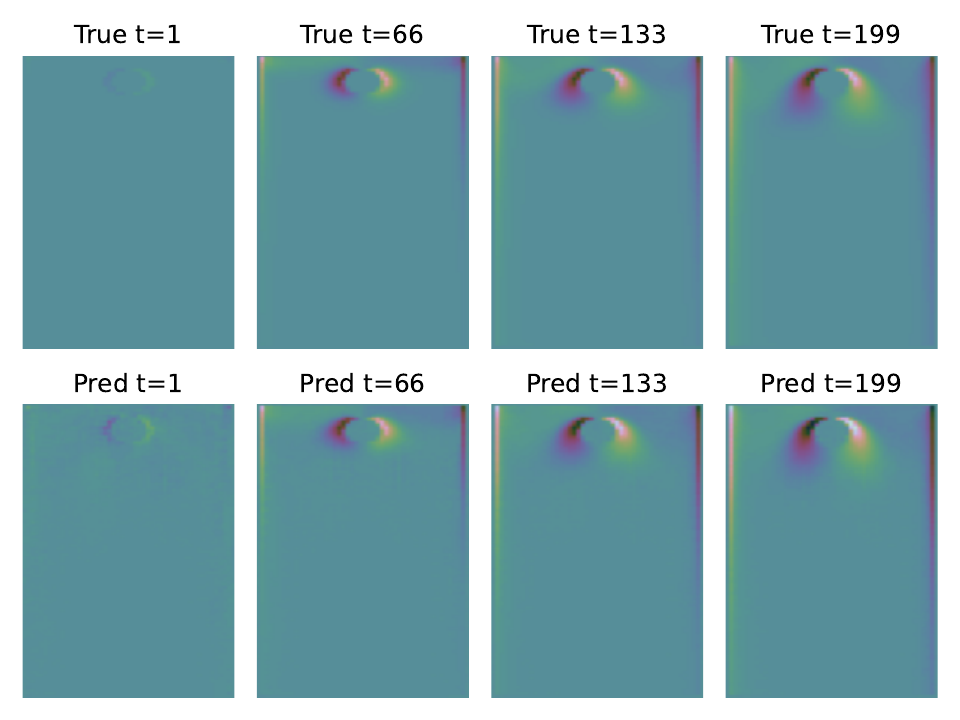}
    \caption{Predictions obtained using RRAEDy on the fluid flow behind a cylinder dataset for a flow with low Reynold's number for different times (test set, only initial condition is given as input).}
    \label{fig:cyl_pred}
\end{figure}

\begin{figure}
    \centering
    \includegraphics[width=0.8\textwidth]{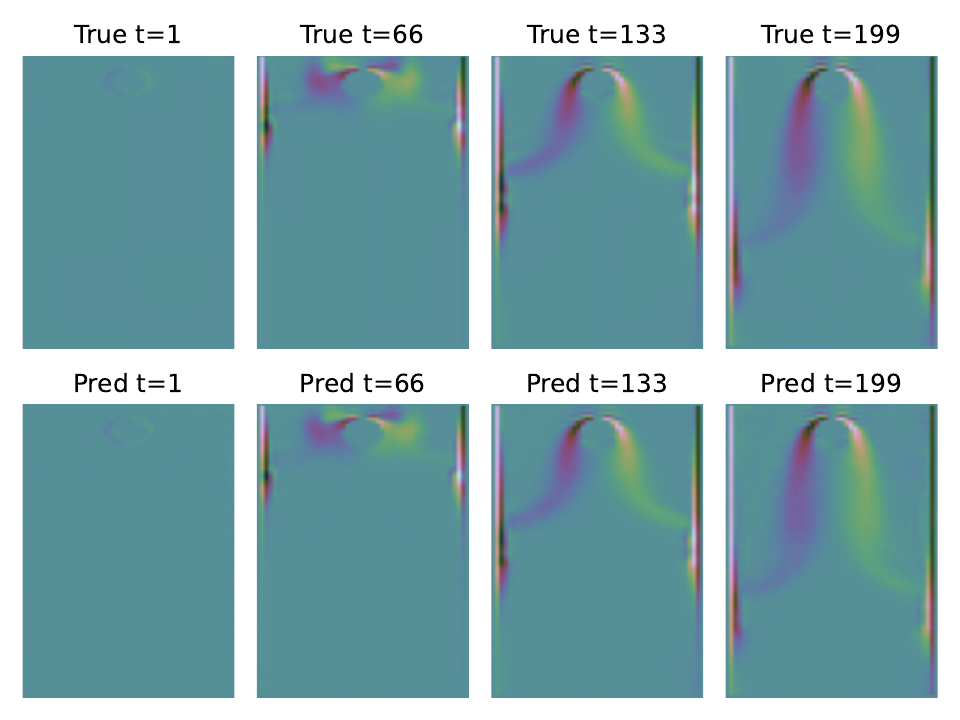}
    \caption{Predictions obtained using RRAEDy on the fluid flow behind a cylinder dataset for a flow with high Reynold's number for different times (test set, only initial condition is given as input).}
    \label{fig:cyl_pred_t}
\end{figure}

\begin{figure}
    \centering
    \includegraphics[width=1\textwidth]{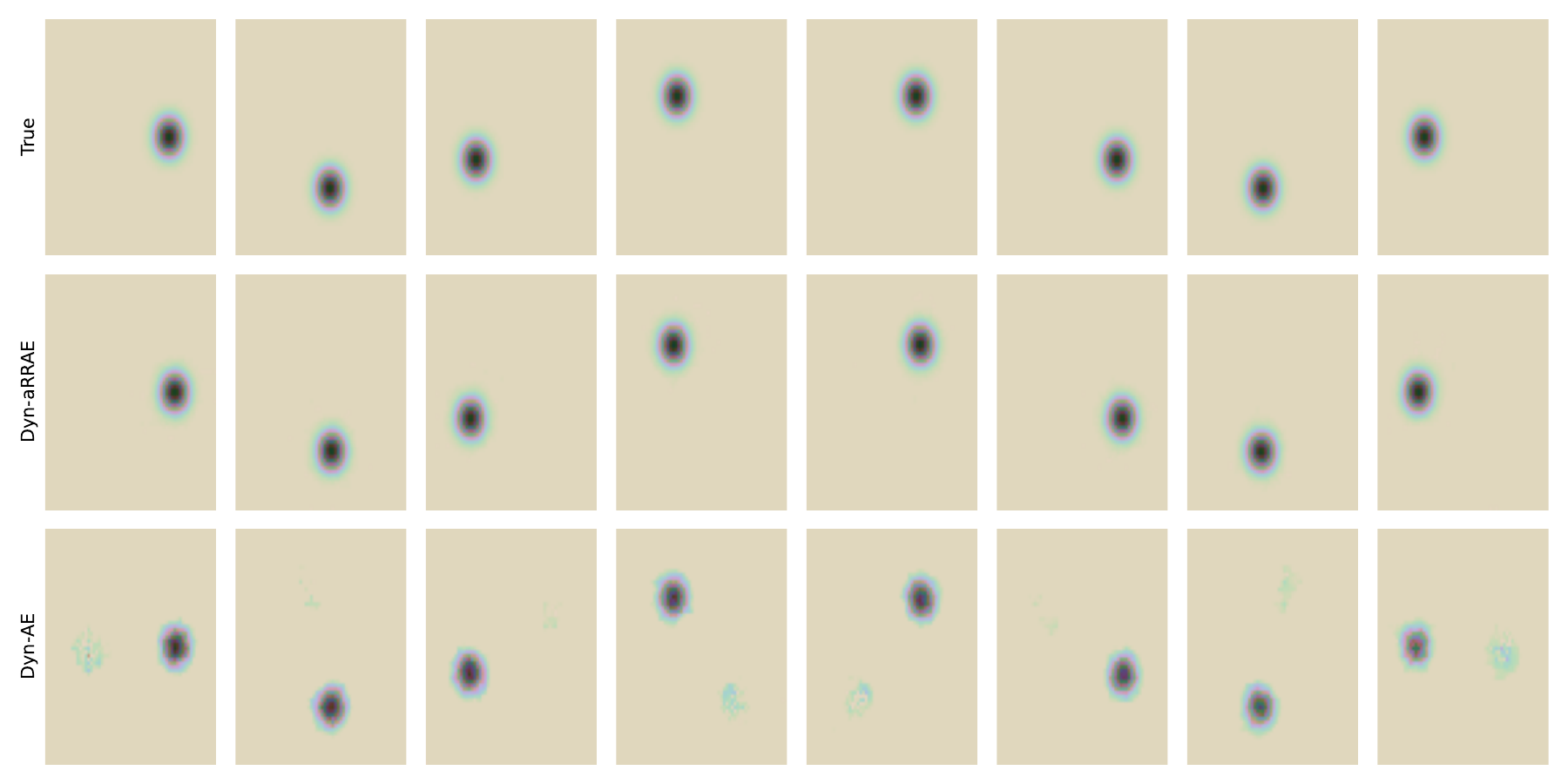}
    \caption{Predictions obtained using RRAEDy on the gaussian problem for $8$ different times to illustrate the circular motion (test set, only initial condition is given as input).}
    \label{fig:gauss_pred_t}
\end{figure}

\begin{figure}
    \centering
    \includegraphics[width=1\textwidth]{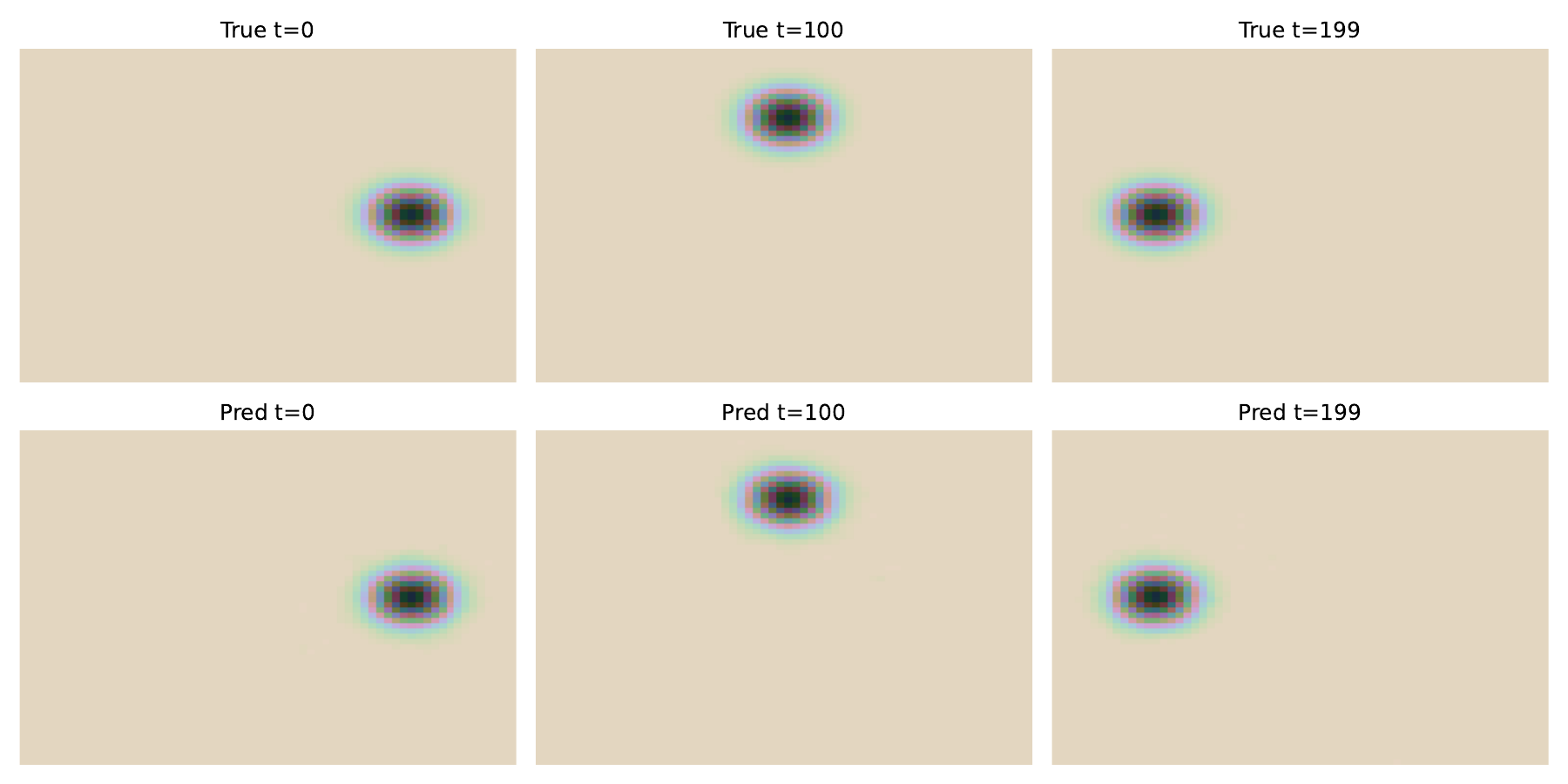}
    \caption{Predictions obtained using RRAEDy on the gaussian problem for $3$ different times to illustrate the reconstruction accuracy (test set, only initial condition is given as input).}
    \label{fig:gauss_pred_t3}
\end{figure}

Next, we present predictions obtained using RRAEDy on the fluid flow behind a cylinder dataset for two flows with low and high Reynold's numbers in Figures \ref{fig:cyl_pred}, and \ref{fig:cyl_pred_t} respectively. The model is able to accurately capture the vortex shedding and complex flow patterns behind the cylinder.

Finally, we present predictions obtained using RRAEDy on the rotation of a 2D Gaussian dataset in Figures \ref{fig:gauss_pred_t} and \ref{fig:gauss_pred_t3}. The model is able to accurately capture the rotation of the Gaussian over time, while maintaining the shape and intensity of the Gaussian.

To evaluate the performance of RRAEDy quantitatively, we compute the norm percentage error (NPE) between the predicted time series $\tilde{X}$ and the ground truth time series $X$ on both the test and train set as follows,
\begin{equation*}
    \text{NPE} = \frac{\|X - \tilde{X}\|_F}{\|X\|_F} \times 100\%,
\end{equation*}

The error found for each dataset is summarized in Table \ref{tab:errors}.

\begin{table}[h!]
    \centering
    \begin{tabular}{|c|c|c|}
        \hline
        \textbf{Dataset} & \textbf{NPE (train)} & \textbf{NPE (test)} \\
        \hline
        Van der Pol Oscillator & $6.67\%$ &  $7.37\%$\\
        \hline
        Burger's Equation & $4.46\%$ & $4.64\%$\\
        \hline
        Fluid behind a cylinder &  $7.13\%$ & $7.21\%$ \\
        \hline
        Rotation of a 2D Gaussian & $4.99\%$ & $6.14\%$ \\
        \hline
    \end{tabular}
    \caption{Norm Percentage Errors (NPE) obtained using RRAEDy for the four datasets.}
    \label{tab:errors}
\end{table}

\section{Theoretical Validation - Ablation Study} \label{sec:ablation}
In this section, we perform an ablation study to evaluate the impact of different components of RRAEDy on the performance and stability of the model. The conclusions of this section are (i) the adaptive bottleneck is crucial to avoid local minima, (ii) the SVD regularization is necessary to ensure marginal stability of the learned dynamics and avoid ``holes'' in the latent space (i.e. overfitting), and (iii) imposing a DMD operator in the latent space is necessary since the dynamics found by the autoencoder are not necessarily linear by themselves.

We obtain the conclusions above by comparing RRAEDy with three variants, each removing a specific component of the model:
\begin{itemize}
    \item \textbf{f-RRAEDy:} This variant uses a fixed bottleneck size $k^*$ throughout the training process, without any adaptation. This allows us to evaluate the impact of the adaptive bottleneck size on the model's performance.
    \item \textbf{AEDy:} This variant uses a standard autoencoder architecture without any SVD-based regularization in the latent space. This helps us assess the importance of the SVD regularization and asses wether the marginal stability proved above leads to better predictions.
    \item \textbf{aRRAE:} This variant removes the DMD operator from the latent space, effectively making it a standard autoencoder with an adaptive bottleneck size. This helps us asses wether the dynamics found by the autoencoder are by themselves linear or if the DMD operator is necessary to capture the dynamics.
\end{itemize}

\textbf{\underline{Note:}} While the variants above help us in understanding the importance of each component of RRAEDy, both f-RRAEDy and AEDy require prior knowledge of the optimal bottleneck size $k^*$, which is usually not known a priori. Therefore, in practice, these variants would require multiple trainings to fine-tune the value of $k^*$, which is not necessary when using RRAEDy.

\begin{table}[!b]
    \centering
    \begin{tabular}{|c|c|c|c|c|}
        \hline
        \textbf{Dataset} & \textbf{RRAEDy} & \textbf{f-RRAEDy} & \textbf{AEDy} & \textbf{aRRAE} \\
        \hline
        Burger's Equation & $\textbf{4.64\%}$ & $5.08\%$ & $5.07\%$ & $39.49\%$\\
        \hline
        Van der Pol Oscillator & $\textbf{7.37\%}$ &  $7.88\%$ & $7.89\%$ & $96.26\%$\\
        \hline
        Fluid behind a cylinder &  $\textbf{7.21\%}$ & $16.36\%$ & $99.97\%$ & $91.9\%$ \\
        \hline
        Rotation of a 2D Gaussian & $\textbf{6.14\%}$ & $15.49\%$ & $17.58\%$ & $31.54\%$ \\
        \hline
    \end{tabular}
    \caption{Norm Percentage Errors (NPE) obtained using RRAEDy and its variants for the four datasets on the test set.}
    \label{tab:ablation_errors}
\end{table}

First, the errors obtained using each of the variants above on the four datasets are summarized in Table \ref{tab:ablation_errors}. The results show that while some of the alternatives have similar error rates in some scenarios, RRAEDy is the only model to achieve the best performance consistently. Further, we note how RRAEDy significantly outperform the other variants on both the Fluid example and the rotated Gaussian problems, where the dynamics tend to be highly nonlinear or local. This shows that both the adaptive scheme, as well as the SVD bottleneck are necessary.

Qualitatively, the rotation of a 2D Gaussian predicted using both RRAEDy and AEDy can be seen in Figure \ref{fig:ablation_gauss}. Note how AEDy have ``holes'' in the latent space, which can be seen by the distortion of the Gaussian shape and the appearance of artifacts. This is a direct consequence of the lack of regularization in the latent space when using standard autoencoders, which leads to overfitting and poor generalization on the test set.

\begin{figure}
    \centering
    \includegraphics[width=1\textwidth]{paper_pred_circular_gauss_ablation.pdf}
    \caption{Predictions obtained using both RRAEDy and AEDy on the rotation of a 2D Gaussian dataset for $8$ different times (test set, only initial condition is given as input). AEDy has ``holes'' in the latent space, which can be seen by the distortion of the Gaussian shape and the appearance of artifacts.}
    \label{fig:ablation_gauss}
\end{figure}

To further explain these results, we investigate the Singular Values and Eigenvalues of the learned DMD operators during training for each of the variants above.

\begin{figure}[!b]
    \centering
    \includegraphics[width=0.8\textwidth]{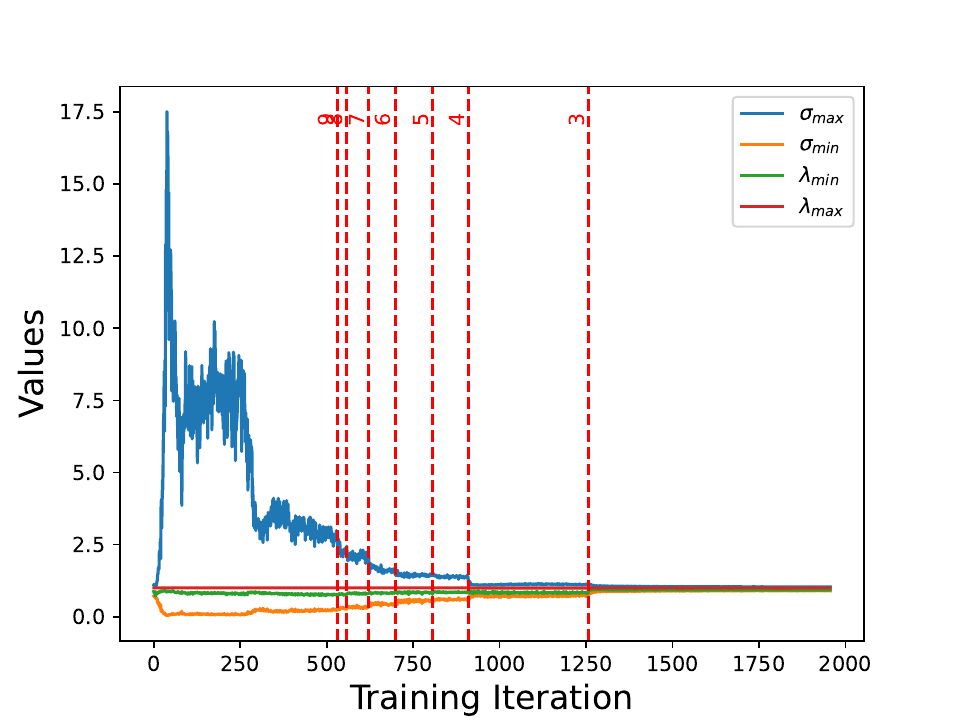}
    \caption{Maximum and Minimum Singular Values and Eigenvalues of the DMD operator $W$ found in the RRAEDy architecture when trained on Burger's equation. The red vertical lines indicate when the value of $k^*$ is decreased by the adaptive algorithm.}
    \label{fig:eig_sing_burger}
\end{figure}

Previously, we have established that both eigenvalues and singular values of the DMD operator $W$ get closer to $1$ as the bottleneck size $k^*$ gets smaller. The biggest/smallest Singular values/ Eigenvalues when training a RRAEDy on burger's equation can be seen in Figure \ref{fig:eig_sing_burger}.

First, note that the red vertical lines are where the value of $k^*$ have been decreased by the adaptive algorithm, starting with a value of $10$. Note that RRAEDy in the beginning of training don't have singular values that are close to $1$. This is mainly because the bound derived in Lemma \ref{lemma:bound} is not tight when $k^*$ is large. Accordingly, RRAEDy can explore some parts where the operator is far from the identity, which is sometimes necessary to avoid converging to local minimums. On the other hand, every time the value of $k^*$ is decreased, both $\sigma_{max}$, and $\sigma_{min}$ get closer to $1$ showing the important effect of $k^*$ on the bound. Therfore, by the end of training, all Singular values and Eigenvalues are close to $1$ which enforces marginal stability of the DMD operator $W$ found by the end of training.

To further investigate empirically how this bound is of advantage to RRAEDy, we study the behavior of the computed DMD operator during training. In what follows, aRRAEs are not included in the study since they do not find a DMD operator during training.

\begin{figure}[!b]
  \centering
  \includegraphics[width=1\textwidth]{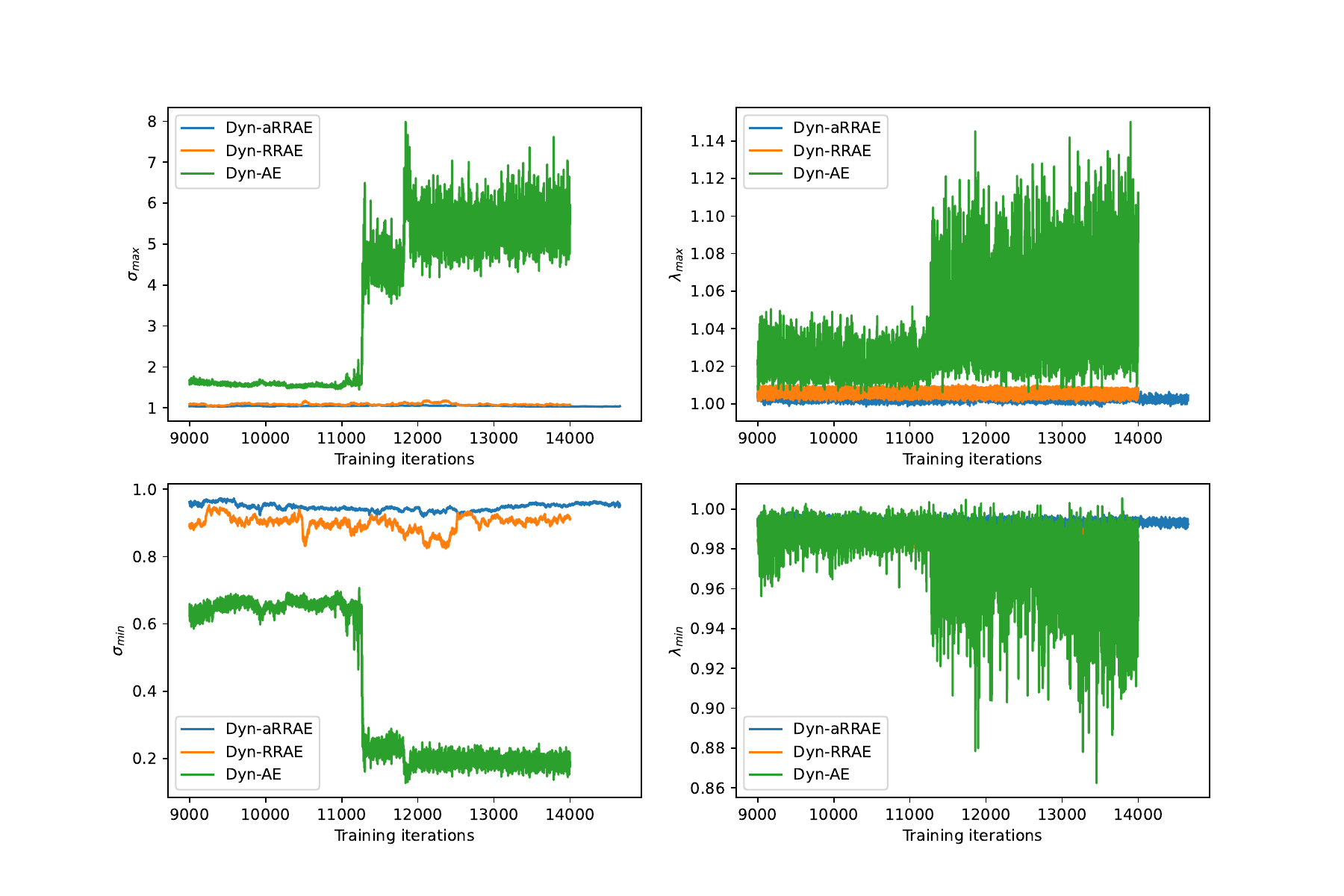}
  \caption{Maximum and Minimum Singular Values and Eigenvalues of the DMD operator for RRAEDy and two variants presented above when trained on the Fluid behind a cylinder problem (last $5000$ forward passes).}
  \label{fig:sing_ablation_fluid}
\end{figure}

The Singular Values, as well as the Eigenvalues of the DMD operator of RRAEDy, f-RRAEDy (without the adaptive) and AEDy (without the SVD), for the last $5000$ forward passes, when trained on the Fluid flow problem,  can be seen in Figure \ref{fig:sing_ablation_fluid}.

\begin{figure}
  \centering
  \includegraphics[width=1\textwidth]{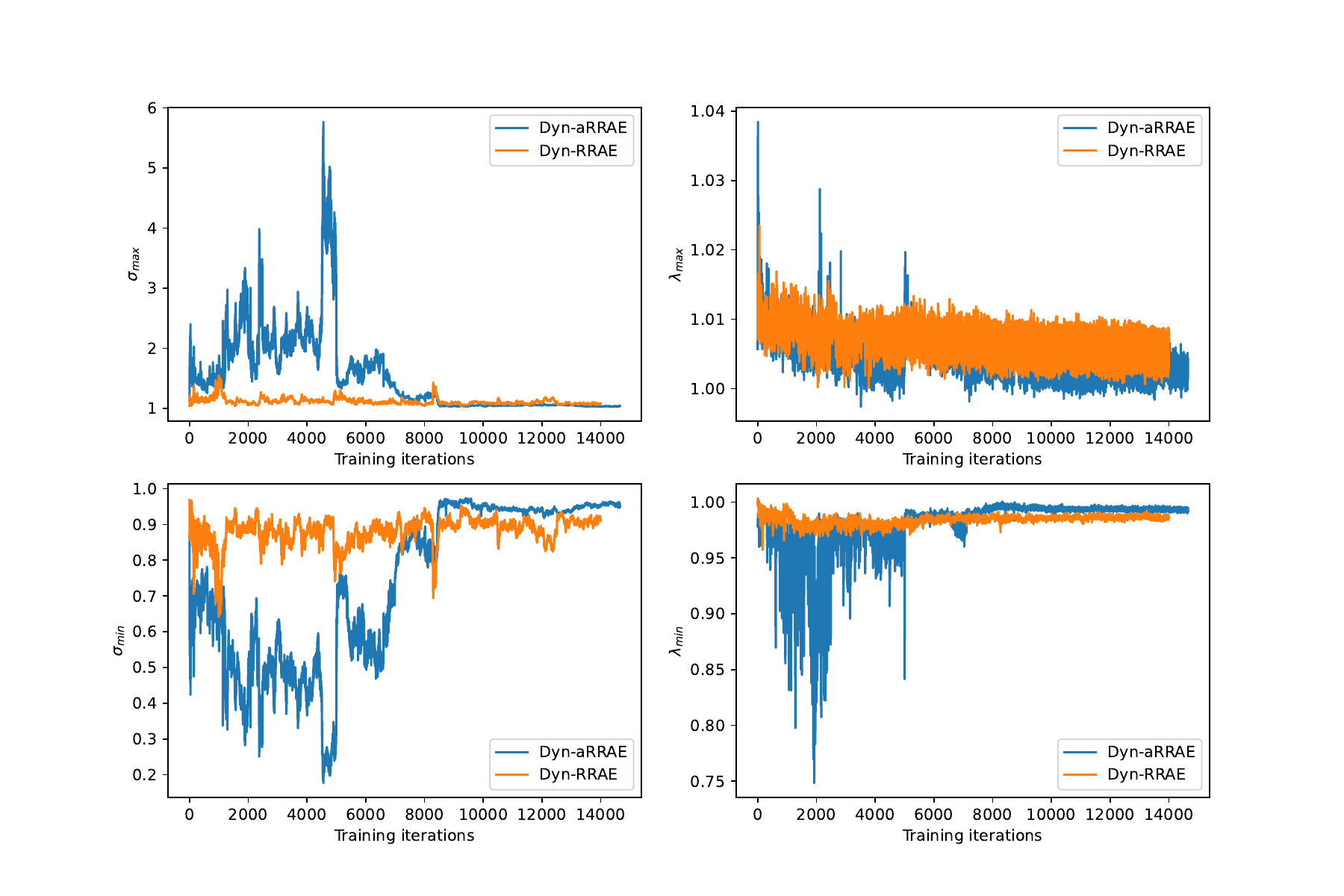}
  \caption{Maximum and Minimum Singular Values and Eigenvalues of the DMD operator of both RRAEDy and f-RRAEDy over the whole training process on the Fluid problem (AEDy are ommitted since they get far from $1$).}
  \label{fig:sing_ablation_fluid_full}
\end{figure}

Note that both singular values and eigenvalues for AEDy (without the SVD) are far from $1$. This leads to highly unstable dynamics, which explains the high error obtained using this variant in Table \ref{tab:ablation_errors}. On the other hand, the singular values and eigenvalues of both RRAEDy and f-RRAEDy over the whole training process when trained on the Fluid problem can be seen in Figure \ref{fig:sing_ablation_fluid_full}.

Note that since $k^*$ is chosen as a fixed small value in f-RRAEDy, the singular values and eigenvalues are close to $1$ from the beginning of training. Contrarily to RRAEDy, f-RRAEDy are not capable of exploring zones where the operator is far from the identity. Accordingly, they can converge to local minimums, as seen in Table \ref{tab:ablation_errors} for the Fluid problem (error of $16.36\%$), or the rotated Gaussian problem (error of $15.49\%$).

Finally, we provide for each final DMD operator $W$ learned using all variants on the four datasets, the norm of $\|W^TW - W^TW\|_2$ to show how close each matrix $W$ is from being normal in Table \ref{tab:commutator_errors}.

\begin{table}[h!]
    \centering
    \begin{tabular}{|c|c|c|c|}
        \hline
        \textbf{Dataset} & \textbf{RRAEDy} & \textbf{f-RRAEDy} & \textbf{AEDy}\\
        \hline
        Burger's Equation & $0.03$ & $0.08$ & $0.27$\\
        \hline
        Van der Pol Oscillator & $0.005$ &  $0.01$ & $0.006$\\
        \hline
        Fluid behind a cylinder &  $0.08$ & $0.05$ & $8.61$ \\
        \hline
        Rotation of a 2D Gaussian & $0.04$ &  $0.009$ & $0.002$\\
        \hline
    \end{tabular}
    \caption{Norm of the commutator $\|W^TW - W^TW\|_2$ for the DMD operator $W$ learned using RRAEDy and its variants for the four datasets on the test set. aRRAEs are ommitted since they do find a DMD operator during training.}
    \label{tab:commutator_errors}
\end{table}

The table above shows that both RRAEDy and f-RRAEDy learn DMD operators that are close to being normal, which is a desirable property for stability. On the other hand, AEDy do not guarantee this property, which can lead to instabilities and worse performance.

\section{ODEs with parameters}
Previously, we have only considered learning time series generated from autonomous ODEs without parameters. In other words, the different initial conditions are enough to determine the state of the system at every time. However, many applications include some parameters $\mu$ that can be modified hence affecting the dynamics of the system. 

\begin{figure}
    \centering
    \includegraphics[width=1\textwidth, trim = 0 8.5cm 0 12.5cm, clip]{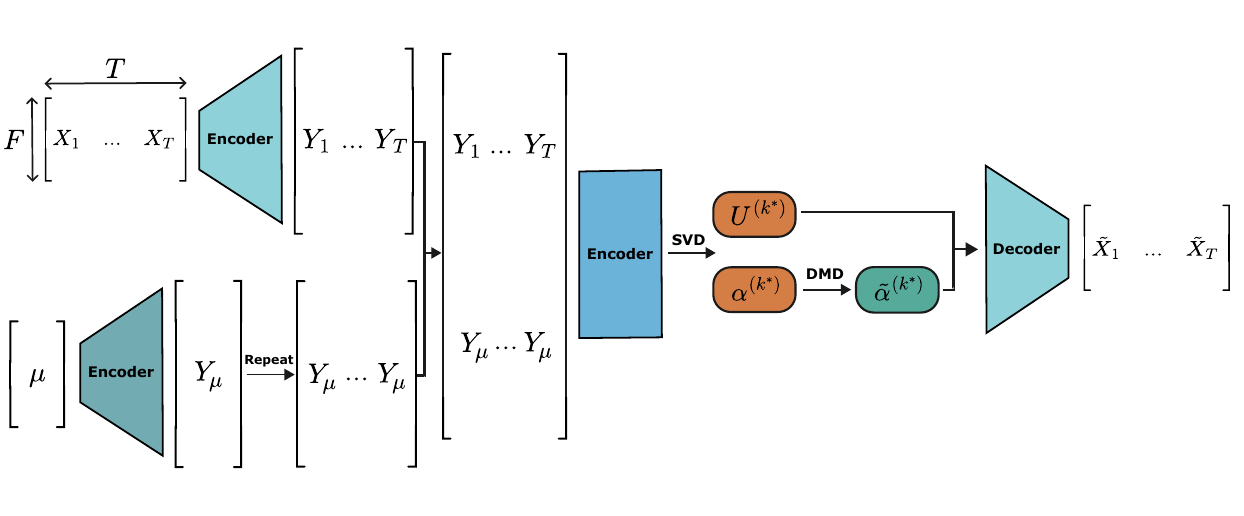}
    \caption{Architecture of RRAEDy (training) with a modified encoder for parametric ODEs when the parameters $\mu$ are constant over time.}
    \label{fig:param_dyn_arrae}
\end{figure}

In this section, we show that RRAEDy can be easily extended to handle more complex forms of ODEs. To do so, we present a simple way of extending RRAEDy to learn time series generated from parametric ODEs of the form,
\begin{equation*}
    \dot{\mathbf{x}} = f(\mathbf{x}, \mu),
\end{equation*}
where $\mu \in \mathbb{R}^P$ is a vector of parameters. To achieve this, we extend our architecture to include the parameters $\mu$ as additional inputs to another encoder as can be seen in Figure \ref{fig:param_dyn_arrae}. The idea is to allow the DMD operator $W$ to depend on the parameters $\mu$, which are encoded and concatenated with the input states.

Note that the architecture presented in Figure \ref{fig:param_dyn_arrae} includes a ``repeat'' operation since it assumes that the parameters $\mu$ are constant over time. The same architecture can be used when the parameters $\mu$ are time-dependent by removing the ``repeat'' operation and providing the parameters at each time step. This architecture is one way to include parameters in RRAEDy, but other architectures that allow $W$ to depend on $\mu$ can also be considered. Note that the theoretical results presented previously still hold for this architecture if the parameters $\mu$ have the same Lipschitz continuity properties as the input states $X$.

In what follows, we present some results obtained using this architecture on the parametric Mass-Spring-Damper system, presented next.

\underline{\textbf{Mass-Spring-Damper System:}} The Mass-Spring-Damper system is a classic mechanical system that consists of a mass attached to a spring and a damper. The system is described by the following second-order differential equation,
\begin{equation*}
    m\frac{d^2x}{dt^2} + c\frac{dx}{dt} + kx = 0,
\end{equation*}
where $m$ is the mass, $c$ is the damping coefficient, and $k$ is the spring constant. The system can be rewritten as a system of first-order ODEs as follows,
\begin{equation*}
    \begin{cases}
        \displaystyle\frac{dx_1}{dt} = x_2,\\[3ex]
        \displaystyle\frac{dx_2}{dt} = -\frac{c}{m}x_2 - \frac{k}{m}x_1.
    \end{cases}
\end{equation*}
For our experiments, we set the initial conditions $x_0=1$, and $v_0=1$. We then generate $N=2000$ samples using random masses, damping coefficients, and spring constants uniformly sampled in the ranges $m\in[0.5, 2.0]$, $c\in[0.1, 2.0]$, and $k\in[0.5, 3.0]$ respectively. Each time series is generated over a time span of $t\in[0, 15]$ with a time step of $\Delta t = 0.05$, resulting in $T=200$ time steps per sample.

\begin{figure}[h!]
    \centering
    \includegraphics[width=0.8\textwidth]{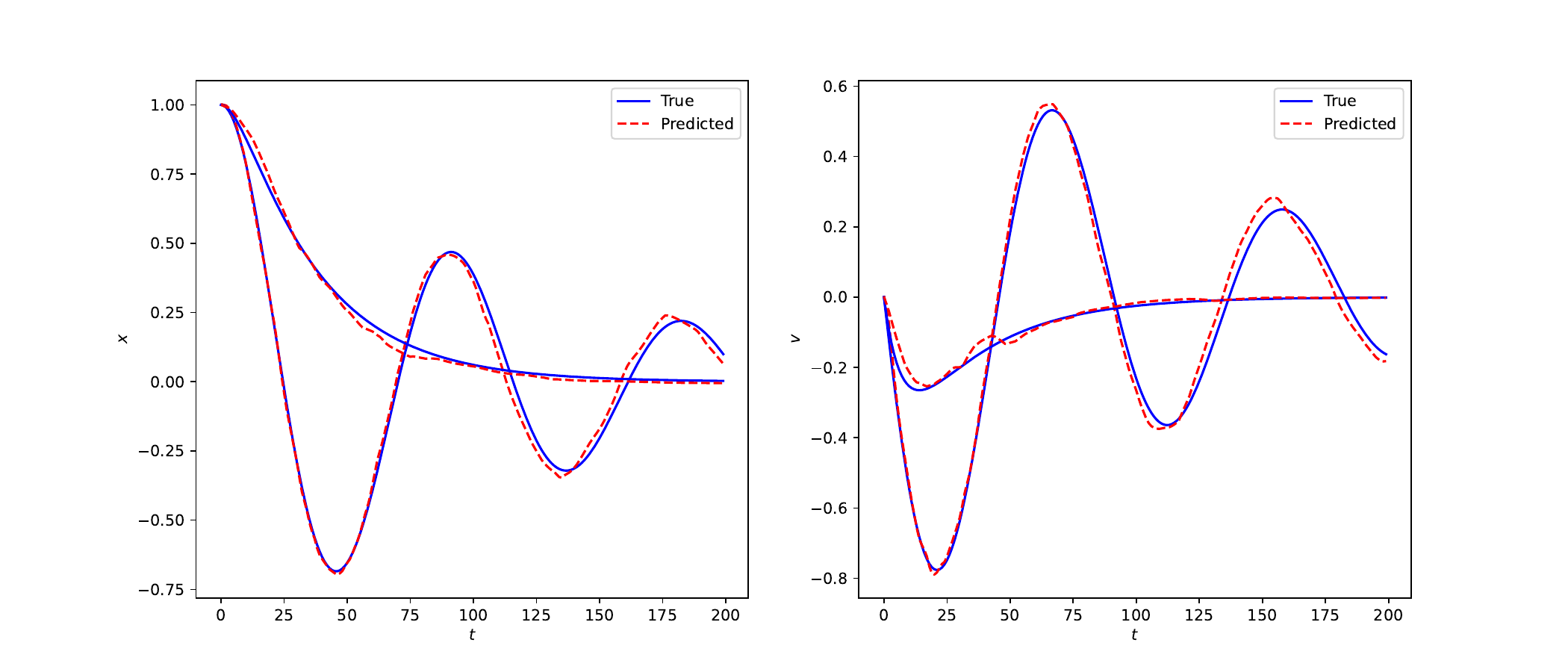}
    \caption{Predictions obtained using RRAEDy on the parametric Mass-Spring-Damper system dataset (test set, only initial condition and parameters are given as input).}
    \label{fig:msd_pred}
\end{figure}

Some predictions obtained using RRAEDy on the parametric Mass-Spring-Damper system can be seen in Figure \ref{fig:msd_pred}. The input of each of the predictions presented is only the initial condition of the test time series along with the parameters, and the model is asked to predict the entire time series. We can see that the model is able to accurately capture the dynamics of the system for different parameter values.

The model achieves a mean squared error (MSE) of $2.9 \times 10^{-3}$ on the test set, demonstrating its ability to generalize across different parameter values. Additionally, the norm of the commutator $\|W^TW - WW^T\|_2$ for the learned DMD operator $W$ is $0.0024$, showing empirically that the theory presented previously still holds for this parametric case.

A video showing a User Interface (UI) where the user can modify the parameters and see the predictions can be found \href{https://youtu.be/ox70mSSMGrM}{here}.

\section{Conclusion and Perspectives}
In this paper, we have introduced Rank Reduction Autoencoders for Dynamical Systems (RRAEDy), a novel architecture for learning stable and accurate representations of time series data generated from dynamical systems. By combining the strengths of autoencoders, Singular Value Decomposition (SVD), and Dynamic Mode Decomposition (DMD), RRAEDy are capable of capturing complex dynamics while ensuring marginal stability through an adaptive bottleneck size.

Experiments on various datasets, including the Van der Pol oscillator, Burger's equation, fluid flow behind a cylinder, and the rotation of a 2D Gaussian, have demonstrated the effectiveness of RRAEDy in accurately predicting time series data. Furthermore, an ablation study highlighted the importance of each component of the architecture, confirming that the adaptive bottleneck, SVD regularization, and DMD operator are all crucial for optimal performance. We also provided an extension of RRAEDy to handle parametric ODEs, showcasing the versatility of the architecture.

Our results show that RRAEDy can converge to better minimums, are robust when trained on data with local behavior, ensure marginal stability of the learned dynamics, and can easily be extended to ODEs with more complex forms.

Future work could explore extensions of RRAEDy to handle more complex scenarios, such as time-dependent parameters, stochastic dynamics, and multi-scale systems. Additionally, applications in real-world datasets could further enhance the utility of RRAEDy in various scientific and engineering domains. Further, some extensions of the DMD operator can be used to capture more complex dynamics in the latent space (e.g., controlled systems, nonlinear dynamics, etc.).

\begin{ack}
This work was supported by the SKF chair in the PIMM laboratory at ENSAM, Paris, France. We also thank the Google TPU Research Cloud program for providing computational resources.
\end{ack}

\newpage
\bibliographystyle{plain}
\bibliography{references}


\appendix
\newpage
\section{Appendix}

\subsection{Proof: Batches converge to similar DMD operators.}\label{ap:proof1}

In this part of the Appendix, we present a proof to the following Lemma,

\newtheorem*{lemma*}{Lemma}

\begin{lemma*}
  When both Hypothesis \ref{hyp:1} and Hypothesis \ref{hyp:2} hold, the learned DMD operators $W_i$ for different batches $i$ converge to similar operators. In other words, we can write,
    \begin{equation}
        W_j \approx Q^TW_iQ, \qquad \text{with,}\qquad QQ^T = Q^TQ = I.
    \end{equation}
\end{lemma*}

\textbf{\underline{Proof:}} Let $i$ and $j$ be two different batches at different epochs containing the same sample. Let $U_i$ and $U_j$ be their corresponding bases found by the SVD in the latent space. Based on Equation \eqref{eq:U_wQ}, we can write $U_j \approx U_iQ$, or $Q \approx U_i^TU_j$. Let $\tilde{X}$ be the reconstruction of the sample, let $\alpha_{DMD}$ be the coefficients predicted by the latent DMD, and let superscripts $batch_i$ and $batch_j$ denote the batch used to compute the quantities. For the same sample passing through different batches at convergence, the predictions using each batch are close, so we can write,

\begin{equation*}
  \begin{aligned}
    \tilde{X}^{batch_i} &\approx \tilde{X}^{batch_j},\\
    \implies \mathcal{D}\left(U_i\alpha^{batch_i}_{DMD}\right) &\approx \mathcal{D}\left(U_j\alpha^{batch_j}_{DMD}\right),\\[2ex]
    \implies U_i\left[\alpha^{batch_i}_{:, 1}, W_i\alpha^{batch_i}_{:, 1}, \dots, W_i^{T-1}\alpha^{batch_i}_{:, 1}\right] &\approx U_j\left[\alpha^{batch_j}_{:, 1}, W_j\alpha^{batch_j}_{:, 1}, \dots, W_j^{T-1}\alpha^{batch_j}_{:, 1}\right].
  \end{aligned}
\end{equation*}
Where we were able to remove the decoder $\mathcal{D}$ from the equation based on Hypothesis \ref{hyp:2}. 

Note that $\alpha^{batch_i}_{:, 1} = U_i^TY_{:, 1}$ and $\alpha^{batch_j}_{:, 1} = U_j^TY_{:, 1}$, where $Y_{:, 1}$ is the encoded initial condition of the sample. In this case, we use $Y_{:, 1}$ for both batches based on Hypothesis \ref{hyp:2}, since the converged encoder generates the same latent space when given the same input. Accordingly, we can write,
\begin{equation*}
  \begin{aligned}
    U_i\left[U_i^TY_{:, 1}, W_iU_i^TY_{:, 1}, \dots, W_i^{T-1}U_i^TY_{:, 1}\right] &\approx U_j\left[U_j^TY_{:, 1}, W_jU_j^TY_{:, 1}, \dots, W_j^{T-1}U_j^TY_{:, 1}\right],\\[2ex]  \implies U_i \mathcal{K}_{T\!-\!1}\!(W_i, U_i^TY_{:, 1}) &\approx U_j\mathcal{K}_{T\!-\!1}\!(W_j, U_j^TY_{:, 1}), \\[2ex]
  \end{aligned}
\end{equation*}
where $\mathcal{K}_{T\!-\!1}\!(W, v) = [v, Wv, \dots, W^{T-1}v]$ is the Krylov matrix of order $T-1$ generated by $W$ and $v$. By multiplying both sides by $U_i^T$, and using Equation \eqref{eq:U_wQ}, we can write,
\begin{equation*}
  \begin{aligned}
    \mathcal{K}_{T\!-\!1}\!(W_i, U_i^TY_{:, 1}) &\approx U_i^TU_j\mathcal{K}_{T\!-\!1}\!(W_j, U_j^TY_{:, 1}),\\[2ex]
    &\approx Q\mathcal{K}_{T\!-\!1}\!(W_j, Q^TU_i^TY_{:, 1}),\\[2ex]
    &\approx \mathcal{K}_{T\!-\!1}\!(QW_jQ^T, U_i^TY_{:, 1}),\\[2ex]
  \end{aligned}
\end{equation*}
where the last approximation is valid since $(QW_jQ^T)^p = QW_j^pQ^T$ for any integer $p$. Since the Krylov matrices are approximately equal, they are based on the same vector $U_i^TY_1$, and based on Hypothesis \ref{hyp:1}, there are at least $k^*$ linearly independent columns in the Krylov matrices (so the matrices are of full rank), we can write\footnote{Detailed proof for this part in Appendix \ref{ap:kyrlov}.},
\begin{equation*}
    W_i \approx QW_jQ^T, \qquad \text{with,}\qquad QQ^T = Q^TQ = I. \eqno\qed
\end{equation*}

\subsection{Krylov Subspace Approximation:}\label{ap:kyrlov}
In this part of the Appendix, we prove the following Lemma,
\newtheorem*{lemmal}{Lemma}
\begin{lemmal}
    Let $v \in \mathbb{R}^{k^*}$, let $W_1, W_2 \in \mathbb{R}^{k^* \times k^*}$, and let the Krylov matrices be defined as,
    \begin{equation*}
        \mathcal{K}_T(W_i, v) = \begin{bmatrix}
            v & W_i v & W_i^2 v & \cdots & W_i^{T-1} v
        \end{bmatrix} \in \mathbb{R}^{k^* \times T}.
    \end{equation*}
    If $\mathcal{K}_T(W_1, v) \approx \mathcal{K}_T(W_2, v)$, they both have full rank, and $T \gg k^*$, then $W_1 \approx W_2$.
\end{lemmal}
\underline{\textbf{Proof:}} First, note that since both $\mathcal{K}_T(W_1, v)$ and $\mathcal{K}_T(W_2, v)$ are full rank, then there are $k^*$ linearly independent columns in each of them. Further, note that since each column of $\mathcal{K}_T(W_i, v)$ is obtained by multiplying the previous column by $W_i$, then the $k^*$ first columns of each matrix are linearly independent. Therefore, for each Krylov matrix, we can write,
\begin{equation*}
    W_i \mathcal{K}_T(W_i, v)  = \mathcal{K}_T(W_i, v) C,
\end{equation*}
with,
\begin{equation*}
C =
\begin{tikzpicture}[baseline=(m.center)]
\node (m) {
$\begin{bmatrix} 0 & 0 & \cdots & 0 & c & \cdots & c\\ 1 & 0 & \vdots & \vdots & c & \cdots & c\\ 0 & 1 & 0 & \vdots & \vdots & \cdots & \vdots\\ \vdots & 0 & \ddots & 0 & \vdots & \cdots & \vdots\\ \vdots & \vdots & 0 & 1 & c & \cdots & c\\ \vdots & \vdots & \vdots & 0 & 0 & 0 & 0\\ \vdots & \vdots & \vdots & \vdots & \vdots & \vdots & \vdots\\ 0 & 0 & 0 & 0 & 0 & 0 & 0\\ \end{bmatrix}$};

\draw[<->, thick, red] ([xshift=0.23cm,yshift=0.2cm]m.north west) -- ([xshift=-2.42cm,yshift=0.2cm]m.north east)
node[midway, above] {$k^*-1$};

\draw[<->, thick, blue] ([xshift=0.1cm,yshift=-0.12cm]m.north east) -- ([xshift=0.1cm,yshift=-3cm]m.north east)
node[midway, right] {$k^*$};
\end{tikzpicture}
\in \mathbb{R}^{T \times T},
\end{equation*}

where $c$ are the coefficients that determine the dependence of the $k^*+1$ column on the previous $k^*$ columns. Note that the matrix $C$ does not depend on $W_i$ since it is constructed solely from the Krylov vectors. Accordingly, we can write,
\begin{equation*}
    W_1 \mathcal{K}_T(W_1, v) = \mathcal{K}_T(W_1, v) C, \quad W_2 \mathcal{K}_T(W_2, v) = \mathcal{K}_T(W_2, v) C.
\end{equation*}
Therefore, since $\mathcal{K}_T(W_1, v) \approx \mathcal{K}_T(W_2, v)$, we can write,
\begin{equation*}
    W_1 = \mathcal{K}_T(W_1, v) C \mathcal{K}_T(W_1, v)^{\dagger} \approx \mathcal{K}_T(W_2, v) C \mathcal{K}_T(W_2, v)^{\dagger} = W_2. \eqno\qed
\end{equation*}

\subsection{Proof: Convergence to Marginally Stable DMD operators.}
\label{ap:proof2}

In this part of the Appendix, we present a proof to the following Lemma,
\newtheorem*{lemmad}{Lemma}

\begin{lemmad}
    Let Hypothesis \ref{hyp:1} and Hypothesis \ref{hyp:3} hold. Let $\kappa\left(Y^{(k^*)}\right)$, and $\sigma_{k^*}$ be the condition number, and the smallest singular value of the truncated latent space $Y^{(k^*)}$ respectively. The eigenvalues of the learned DMD operator $W$ satisfy,
    \begin{equation*}
        |\lambda_i -1| \leq \frac{\sqrt{k^*}}{\sigma_{k^*}}L_EL_X, \quad \forall i\in[1, k^*].
    \end{equation*}
    Further, we can bound the norm of $W - I$ as follows,
    \begin{equation*}
        \|W - I\|_2 \leq \frac{\sqrt{k^*}}{\sigma_{k^*}}\kappa\left(Y^{k^*}\right)L_EL_X.
    \end{equation*}

\end{lemmad}

\textbf{\underline{Proof:}} First, let $Y^{(k^*)} = U\Sigma V^T = U\Sigma\Omega$ be the SVD of the truncated latent space (we ommit the superscript $(k^*)$ for simplicity).

In what follows, we first prove marginal stability for the DMD operator $W_{V^T}$, computed using the right singular vectors of $Y^{(k^*)}$, and then we extend the proof to $W$, which is computed using the coefficients $\Sigma V^T = \Sigma \Omega$.

Note that based on Hypothesis \ref{hyp:3}, we can bound the difference between two consecutive columns of $\Omega$ as follows,
\begin{equation*}
  \begin{aligned}
    \epsilon_t := \|\Omega_{:, t+1} - \Omega_{:, t}\|_2 &= \|\Sigma^{-1}U^T\left(Y_{:, t+1} - Y_{:, t}\right)\|_2 \\
    &\leq \|\Sigma^{-1}\|_2\|U^T\|_2\|Y_{:, t+1} - Y_{:, t}\|_2 \\
    &\leq \frac{L_EL_X}{\sigma_{k^*}}, \quad \forall t\in[1, T-1].
  \end{aligned}
\end{equation*}
This is equivalent to writing,
\begin{equation}\label{eq:V_diff}
    \Omega_{:, t+1} = \Omega_{:, t} + \underline{\epsilon}_t, \qquad \text{with,} \qquad \|\underline{\epsilon}_t\|_2 = \epsilon_t \leq \frac{L_EL_X}{\sigma_{k^*}}, \qquad  \forall t\in[1, T-1].
\end{equation}
Now, let $\Omega_p \in\mathbb{R}^{k^* \times (T-1)}$ and $\Omega_m \in\mathbb{R}^{k^* \times (T-1)}$ be defined as follows,
\begin{equation*}
    \Omega_p = \Omega_{:, 2:T} = [\Omega_{:, 2}, \dots, \Omega_{:, T}], \qquad \Omega_m = \Omega_{:, 1:T-1} = [\Omega_{:, 1}, \dots, \Omega_{:, T-1}].
\end{equation*}
The DMD operator $W_{V^T} = W_{\Omega}$ computed using $\Omega$ can be written as,
\begin{equation}\label{eq:W_v}
    W_\Omega = \Omega_p\left(\Omega_m\right)^{\dagger} = \underbrace{\Omega_p\Omega_m^T}_{:=G_{p,m}}\left(\underbrace{\Omega_m\Omega_m^T}_{:=G_{m, m}}\right)^{-1} = G_{p,m}G_{m, m}^{-1}.
\end{equation}
Note that since $\Omega = V^T$, and $V^TV = I$, each entry of $G_{m,m}$ can be written as,
\begin{equation*}
    \left(G_{m, m}\right)_{i, j} = \sum_{t=1}^{T-1} \Omega_{i, t} \Omega_{t, j} = \begin{cases}
        1 - \Omega_{i, T}\Omega_{T, j}, & \text{if } i=j,\\
        0 - \Omega_{i, T}\Omega_{T, j}, & \text{if } i\neq j.
    \end{cases}
\end{equation*}
Accordingly, we can write,
\begin{equation}\label{eq:Gmm}
    G_{m, m} = I - \Omega_{:, T}\Omega_{T, :}, \quad \text{similarly,} \quad G_{p, p} = I - \Omega_{:, 1}\Omega_{1, :}.
\end{equation}
Now, based on Equation \eqref{eq:V_diff} we can express $G_{p, m}$ as follows,
\begin{equation*}
  G_{p, m} = \sum_{t=1}^{T-1} \Omega_{:, t+1}\Omega_{t, :} = \sum_{t=1}^{T-1} \left(\Omega_{:, t} + \underline{\epsilon}_t\right)\Omega_{t, :} = G_{m, m} + \sum_{t=1}^{T-1}\underline{\epsilon}_t\Omega_{t, :}\,\,.
\end{equation*}
Replacing these expressions in the expression of $W_{V^T}=W_\Omega$ in Equation \eqref{eq:W_v}, we get,
\begin{equation*}
  \begin{aligned}
    W_\Omega &= \left(G_{m, m} + \sum_{t=1}^{T-1}\underline{\epsilon}_t\Omega_{t, :}\right)G_{m, m}^{-1},\\[2ex]
    &= I + \left(\sum_{t=1}^{T-1}\underline{\epsilon}_t\Omega_{t, :}\right)G_{m, m}^{-1},\\[2ex]
    &= I + EG_{m, m}^{-1}, \quad \text{with,} \quad E = \sum_{t=1}^{T-1}\underline{\epsilon}_t\Omega_{t, :}\,\,.
  \end{aligned}
\end{equation*}
Accordingly, we can bound the norm of $W_\Omega - I$ as follows,
\begin{equation*}
    \|W_\Omega - I\|_2 \leq \|E\|_2\|G_{m, m}^{-1}\|_2,\\[2ex]
\end{equation*}
In what follows, we bound both terms on the right-hand side. First, we have,
\begin{equation*}
  \begin{aligned}
    \|E\|_2 = \left\|\sum_{t=1}^{T-1}\underline{\epsilon}_t\Omega_{t, :}\right\|_2 &\leq \sum_{t=1}^{T-1}\|\underline{\epsilon}_t\Omega_{t, :}\|_2,\\[2ex]
    &\leq \sum_{t=1}^{T-1}\|\underline{\epsilon}_t\|_2\|\Omega_{t, :}\|_2,\\[2ex]
    \text{(Cauchy-Schwarz)} \quad &\leq \left(\
    \sum_{t=1}^{T-1}\|\underline{\epsilon}_t\|_2^2\right)^{1/2} \left(\sum_{t=1}^{T-1}\|\Omega_{t, :}\|_2^2\right)^{1/2},\\[2ex]
    \text{(i.i.d. entries with Lipschitz Encoder)} \quad &\leq \sqrt{(T-1)}\max_t(\epsilon_t)\sqrt{\frac{(T-1)k}{T}},\\[2ex]
    \left((T-1) < T\right) \qquad &\leq \frac{\sqrt{k^*}L_EL_X}{\sigma_{k^*}}.
  \end{aligned}
\end{equation*}
Second, we can find $\|G_{m, m}^{-1}\|_2$ as follows,
\begin{equation*}
    \|G_{m, m}^{-1}\|_2 = \frac{1}{\sigma_{min}(G_{m, m})}.
\end{equation*}
Note from Equation \eqref{eq:Gmm} that $G_{m, m}$ is a Gram matrix, so it is symmetric, and its singular values are the absolute values of its eigenvalues. Accordingly, since $k^* \ll T$, we can write,
\begin{equation*}
    \sigma_{min}(G_{m, m}) = \min\left(1, 1 - \|\Omega_{T, :}\|_2^2\right) = 1 - \|\Omega_{T, :}\|_2^2 \approx 1 - \frac{k^*}{T},
\end{equation*}
since we assumed i.i.d entries and the encoder is Lipschitz continuous (Hypotheses \ref{hyp:1}, and \ref{hyp:3}). Accordingly, we can write,
\begin{equation*}
    \|G_{m, m}^{-1}\|_2 \approx \frac{1}{1 - \frac{k^*}{T}} \approx 1.
\end{equation*}
Using both of these bounds and plugging them back into Equation \eqref{eq:W_v}, we get,
\begin{equation}\label{eq:WvmI}
    \|W_\Omega - I\|_2 \leq \frac{\sqrt{k^*}L_EL_X}{\sigma_{k^*}}.
\end{equation}
By the Bauer-Fike theorem, the eigenvalues of $W_\Omega$ satisfy,
\begin{equation*}
    |\lambda_i -1| \leq \|W_\Omega - I\|_2 \leq \frac{\sqrt{k^*}L_EL_X}{\sigma_{k^*}}, \quad \forall i\in[1, k^*].
\end{equation*}
Note that $W$ and $W_\Omega$ are related as follows,
\begin{equation*}
    W = \Sigma W_\Omega \Sigma^{-1},
\end{equation*}
since $\alpha^{(k^*)} = \Sigma V^T = \Sigma \Omega$. Accordingly, $W$, and $W_\Omega$ are similar matrices and they have the same eigenvalues, which concludes the first part of proof.

Finally we can write,
\begin{equation*}
  \|W - I\|_2 = \|\Sigma W_\Omega\Sigma^{-1} - \Sigma I \Sigma^{-1}\|_2 \leq \|\Sigma\|_2\|W_\Omega - I\|_2\|\Sigma^{-1}\|_2 = \kappa\left(Y^{(k^*)}\right)\|W_\Omega - I\|_2.
\end{equation*}
By replacing the bound of $\|W_\Omega - I\|_2$ from Equation \eqref{eq:WvmI}, we conclude the proof. \hfill \qed

\subsection{RRAEDy when doing batches:}\label{ap:batches}
In this part of the Appendix, we present the RRAEDy model when multiple samples are available. The model remains similar to the architecture presented in the paper but with a few extensions. Let $X\in\mathbb{R}^{F\times T\times N}$ denote the data tensor containing $N$ time series (or samples) of length $T$, where each observation is a vector of size $F$.

First, for a tensor $\mathcal{T}\in\mathbb{R}^{F\times T\times N}$, we define the following operations,
\begin{equation*}
  \begin{cases}
    \text{Flatten}(\mathcal{T}) = \begin{bmatrix}
       & & & & & & & &\\
      \mathcal{T}_{:, 1, 1} & \mathcal{T}_{:, 2, 1} & \cdots & \mathcal{T}_{:, T, 1} & \mathcal{T}_{:, 1, 2} & \cdots & \mathcal{T}_{:, T, 2} & \cdots & \mathcal{T}_{:, T, N} \\
       & & & & & & & &\\
    \end{bmatrix}, \\[4ex]
    \text{Unflatten}(\text{Flatten}(\mathcal{T})) = \mathcal{T}.\\[1ex]
  \end{cases}
\end{equation*}
Note that Flatten($\mathcal{T}$) $\in\mathbb{R}^{F \times (T . N)}$ is a matrix where each column is an observation of the time series. The model is defined as follows,
\begin{equation*}
  \begin{cases}
    \text{\underline{Encoding:}}\\[2ex]
    Y_{:, j, i}= \mathcal{E}(X_{:, j, i}), \quad \forall j\in[1, T], \, \forall i\in[1, N],  &\text{with,} \quad Y \in \mathbb{R}^{L\times T\times N}, \qquad \mathcal{E}: \mathbb{R}^{F} \xrightarrow{} \mathbb{R}^{L},\\[2ex]
    \hat{Y} = \text{Flatten}(Y), &\text{with}, \quad \hat{Y}\in\mathbb{R}^{L \times (T . N)}, \\[2ex]
    \text{\underline{Latent SVD:}}\\[2ex]
    \hat{Y}^{(k^*)} = \displaystyle\underbrace{\sum_{i=1}^{k^*}\hat{U}_{:, i}\hat{\sigma}_i\hat{V}_{:, i}^T}_{\text{Truncated SVD}} = \hat{U}^{(k^*)}\hat{\alpha}^{(k^*)}, &\text{with,} \quad \hat{U}^{(k^*)} \in\mathbb{R}^{L\times k^*}, \quad \hat{\alpha}^{(k^*)} \in\mathbb{R}^{k^* \times (T .N)}, \\[8ex]
    \alpha^{(k^*)} = \text{Unflatten}(\hat{\alpha}^{(k^*)}),  &\text{with,} \quad \alpha^{(k^*)}\in\mathbb{R}^{k^*\times T \times N},\\[2ex]
    \text{\underline{Latent DMD:}}\\[2ex]
    \multicolumn{2}{l}{\hspace{-0.15cm}\alpha^{(k^*)}_m = \text{Flatten}\left(\alpha_{:, 1:T-1, :}^{(k^*)}\right), \qquad \alpha^{(k^*)}_p = \text{Flatten}\left(\alpha_{:, 2:T, :}^{(k^*)}\right),} \\[2ex]
    W = \alpha^{(k^*)}_p \left(\alpha^{(k^*)}_m\right)^{\dagger}, & \text{with,}\quad W\in\mathbb{R}^{k^*\times k^*},\\[2ex]
    \tilde{\alpha}^{(k^*)}_{:, :, i} = \left[\alpha^{(k^*)}_{:, 1, i}, W\alpha^{(k^*)}_{:, 1, i}, \dots, W^{T-1}\alpha^{(k^*)}_{:, 1, i}\right], \quad \forall i\in[1, N],\\[2ex]
    \text{\underline{Latent Reconstruction:}}\\[2ex]
    \tilde{Y}^{(k^*)}_{:, :, i} = U^{(k^*)}\tilde{\alpha}^{(k^*)}_{:, :, i}, \quad \forall i\in[1, N],\\[2ex]
    \text{\underline{Decoding:}}\\[2ex]
    \tilde{X}_{:, j, i}= \mathcal{D}(Y^{k^*}_{:, j, i}) \quad \forall j\in[1, T], \, \forall i\in[1, N],  &\text{with,} \quad \tilde{X} \in \mathbb{R}^{F\times T \times N}, \qquad \mathcal{D}: \mathbb{R}^{L} \xrightarrow{} \mathbb{R}^{F}.\\[1ex]
  \end{cases}
\end{equation*}

The only difference with the architecture presented in the main paper is the use of the Flatten and Unflatten operations to handle multiple samples by the SVD and DMD. However, since during inference, both the SVD basis $U_f$ and the DMD operator $W$ are fixed, the inference process remains the same as in the single sample case, i.e., each sample can be inferred independently.

\subsection{Model Architecture and Training:}\label{ap:architecture}
In what follows, we provide details about the architecture and training hyperparameters used for the experiments presented in the main paper. For more details about the implementation, please refer to the code repository, available \href{https://github.com/JadM133/RRAEDy}{here}. First, we begin by presenting the architecture that was used for the 1D problems (i.e., Van der Pol oscillator and Burger's equation). For these problems, we used fully connected neural networks (Multilayer Perceptrons (MLP)) for both the encoder and decoder. The architecture of both the encoder and decoder were fixed for both 1D problems and is summarized in Table \ref{tab:mlp_architecture}.

\renewcommand{\thefigure}{A.3-\arabic{figure}}
\setcounter{figure}{0}
\renewcommand{\thetable}{A.2-\arabic{table}}
\setcounter{table}{0}

\begin{table}[!h]
    \centering
    \begin{tabular}{|c|c|c|}
        \hline
        \textbf{Layer} & \textbf{Encoder} & \textbf{Decoder} \\
        \hline
        Input & $F$ & $L$ \\
        \hline
        Hidden Layer 1 & 64 (ReLU) & 64 (ReLU) \\
        \hline
        Hidden Layer 2 & 64 (ReLU) & 64 (ReLU) \\
        \hline
        Hidden Layer 3 & 64 (ReLU) & 64 (ReLU) \\
        \hline
        Hidden Layer 4 & - & 64 (ReLU) \\
        \hline
        Hidden Layer 5 & - & 64 (ReLU) \\
        \hline
        Hidden Layer 6 & - & 64 (ReLU) \\
        \hline
        Output & $L$ (Linear) & $F$ (Linear) \\
        \hline
    \end{tabular}
    \caption{Architecture of the encoder and decoder used for the 1D problems.}
    \label{tab:mlp_architecture}
\end{table}

On the other hand, for the 2D problems (i.e., fluid flow behind a cylinder and Rotation of a 2D Gaussian), we used Convolutional Neural Networks (CNNs) for both the encoder and decoder with some fully connected layer in between. The architecture of both the encoder and decoder can be found in Table \ref{tab:cnn_architecture}.

\begin{table}[!h]
    \centering
    \begin{tabular}{|c|c|c|}
        \hline
        \textbf{Layer} & \textbf{Encoder} & \textbf{Decoder} \\
        \hline
        Input & $D_1\times D_2$ & $L$ \\
        \hline
        Layer 0 & - &  $L_{CNN}$\\
        \hline
        Layer 1 & - &  Unflatten (32 channels) \\
        \hline
        Layer 1 & 32 filters, $3\times 3$, stride 2 (ReLU) & Tr. Conv, 64 filters, $3\times 3$, stride 2 (ReLU) \\
        \hline
        Layer 2 & 64 filters, $3\times 3$, stride 2 (ReLU) & Tr. Conv, 32 filters, $3\times 3$, stride 2 (ReLU) \\
        \hline
        Layer 3 & 128 filters, $3\times 3$, stride 2 (ReLU) & - \\
        \hline
        Layer 4 & 256 filters, $3\times 3$, stride 2 (ReLU) & - \\
        \hline
        Layer 5 & Flatten & - \\
        \hline
        Output & $L$ (Linear) & 1 filter, $k_{CNN}$, stride 1 (Linear) \\
        \hline
    \end{tabular}
    \caption{Architecture of the encoder and decoder used for the 2D problems.}
    \label{tab:cnn_architecture}
\end{table}
Note that both $L_{CNN}$ and $k_{CNN}$ depend on the input size $D$ and the desired latent size $L$. In our code, first $L_{CNN}$ is computed as the dimension with which the shape of the output is closest to the shape of the input for the given decoder architecture. Then, $k_{CNN}$ is computed to fix the output size of the decoder to be equal to the input size $D_1\times D_2$. The values of $L_{CNN}$ and $k_{CNN}$ used for each of the 2D problems presented in the main paper are summarized in Table \ref{tab:cnn_params}.

\begin{table}[!h]
    \centering
    \begin{tabular}{|c|c|c|}
        \hline
        \textbf{Dataset} & \textbf{$L_{CNN}$} & \textbf{$k_{CNN}$} \\
        \hline
        Fluid behind a cylinder & 10400 & (1, 3) \\
        \hline
        Rotation of a 2D Gaussian & 8192 & (1,1) \\
        \hline
    \end{tabular}
    \caption{Values of $L_{CNN}$ and $k_{CNN}$ used for each of the 2D problems.}
    \label{tab:cnn_params}
\end{table}

The training hyperparameters used for each of the datasets presented in the main paper are summarized in Table \ref{tab:training_hyperparams}.

\begin{table}[!h]
    \centering
    \begin{tabular}{|c|c|c|c|c|c|}
        \hline
        \textbf{Dataset} & \textbf{Batch Size} & \textbf{Learning Rate} & \textbf{Epochs} & \textbf{$L$} & \textbf{$k^*$} \\
        \hline
        Van der Pol Oscillator & 32 & $1\times 10^{-3}$ & 15000 & 10 & 3\\
        \hline
        Burger's Equation & 32 & $1\times 10^{-3}$ & 10000 & 10 & 3\\
        \hline
        Fluid behind a cylinder & 32 & $1\times 10^{-3}$ & 14000 & 64 & 9\\
        \hline
        Rotation of a 2D Gaussian & 32 & $1\times 10^{-3}$ & 16400 & 64 & 4\\
        \hline
    \end{tabular}
    \caption{Training hyperparameters used for each dataset. The value of $k^*$ is the one automatically discovered by the adaptive algorithm.}
    \label{tab:training_hyperparams}
\end{table}

\subsection{Extrapolation using RRAEDy:}\label{ap:extrapolation}

\begin{figure}[!h]
    \centering
    \includegraphics[width=0.6\textwidth]{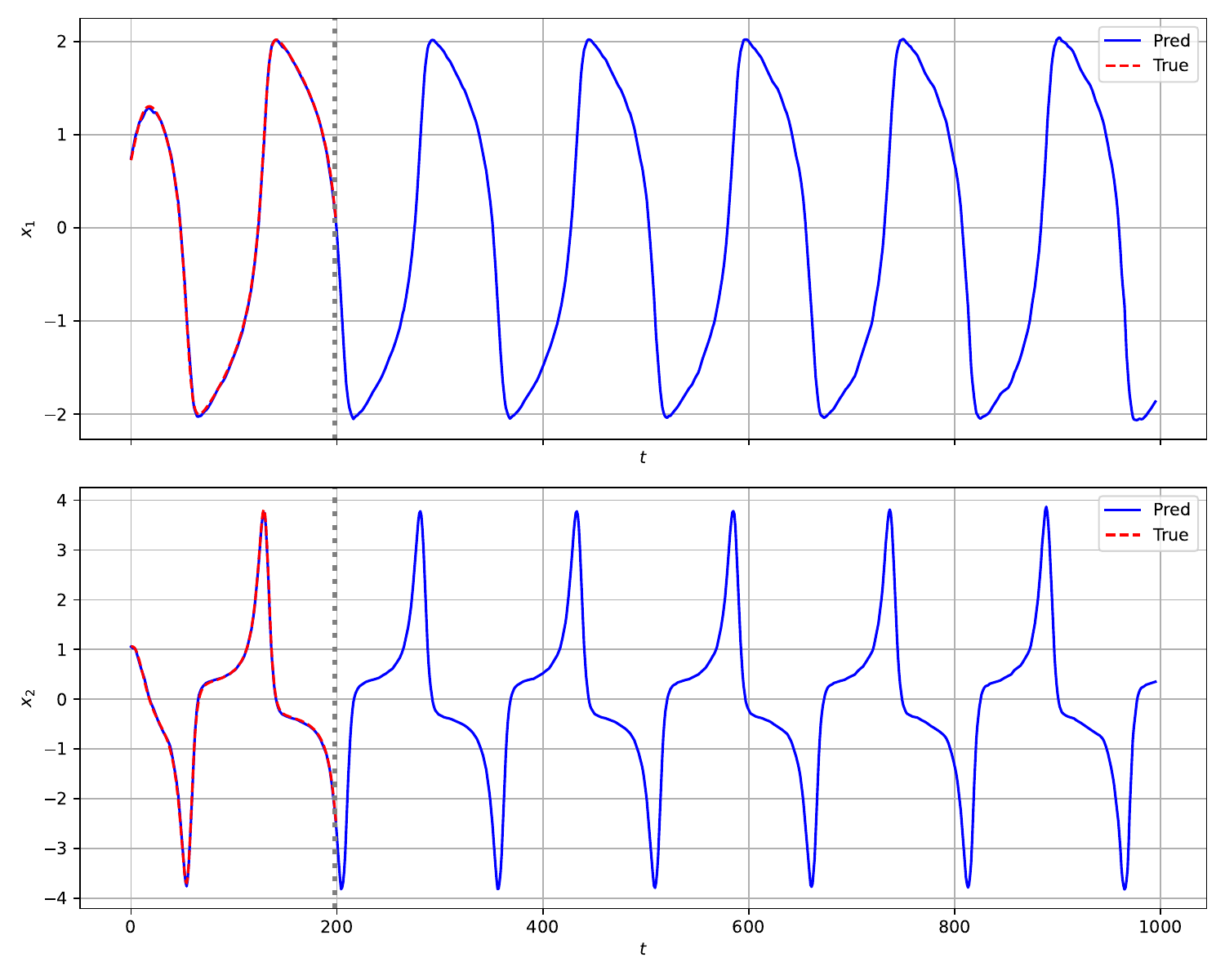}
    \caption{Extrapolation of the learned dynamics using RRAEDy on the Van der Pol oscillator dataset to $5$ times the training time horizon.}
    \label{fig:vdp_extrapolation}
\end{figure}

\begin{figure}[!h]
    \centering
    \includegraphics[width=0.8\textwidth]{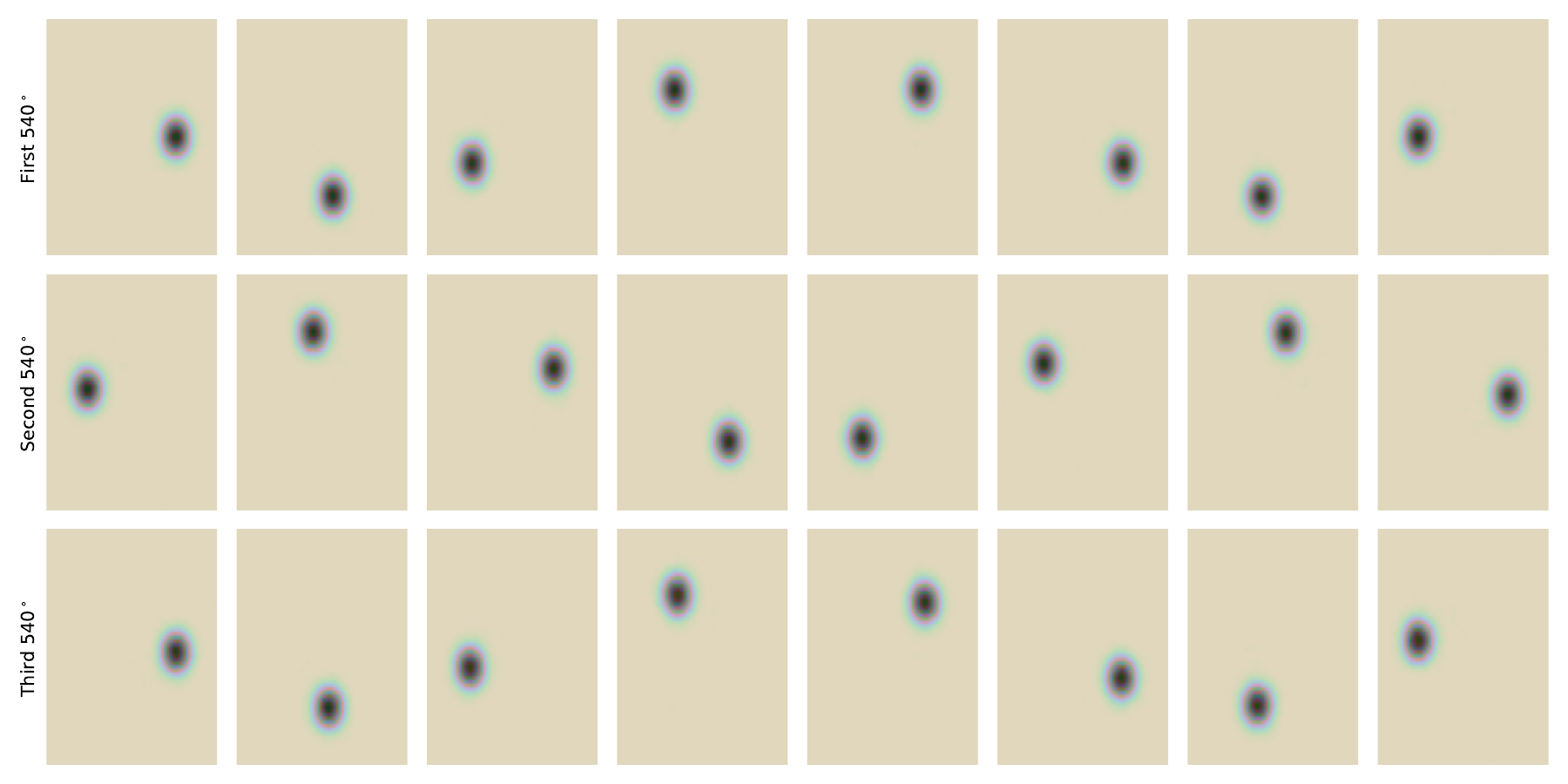}
    \caption{Extrapolation of the learned dynamics using RRAEDy on the Rotation of a 2D Gaussian dataset to $3$ times the training time horizon (first line is training).}
    \label{fig:rotated_gaussian_extrapolation_RR}
\end{figure}

\begin{figure}[!h]
    \centering
    \includegraphics[width=0.8\textwidth]{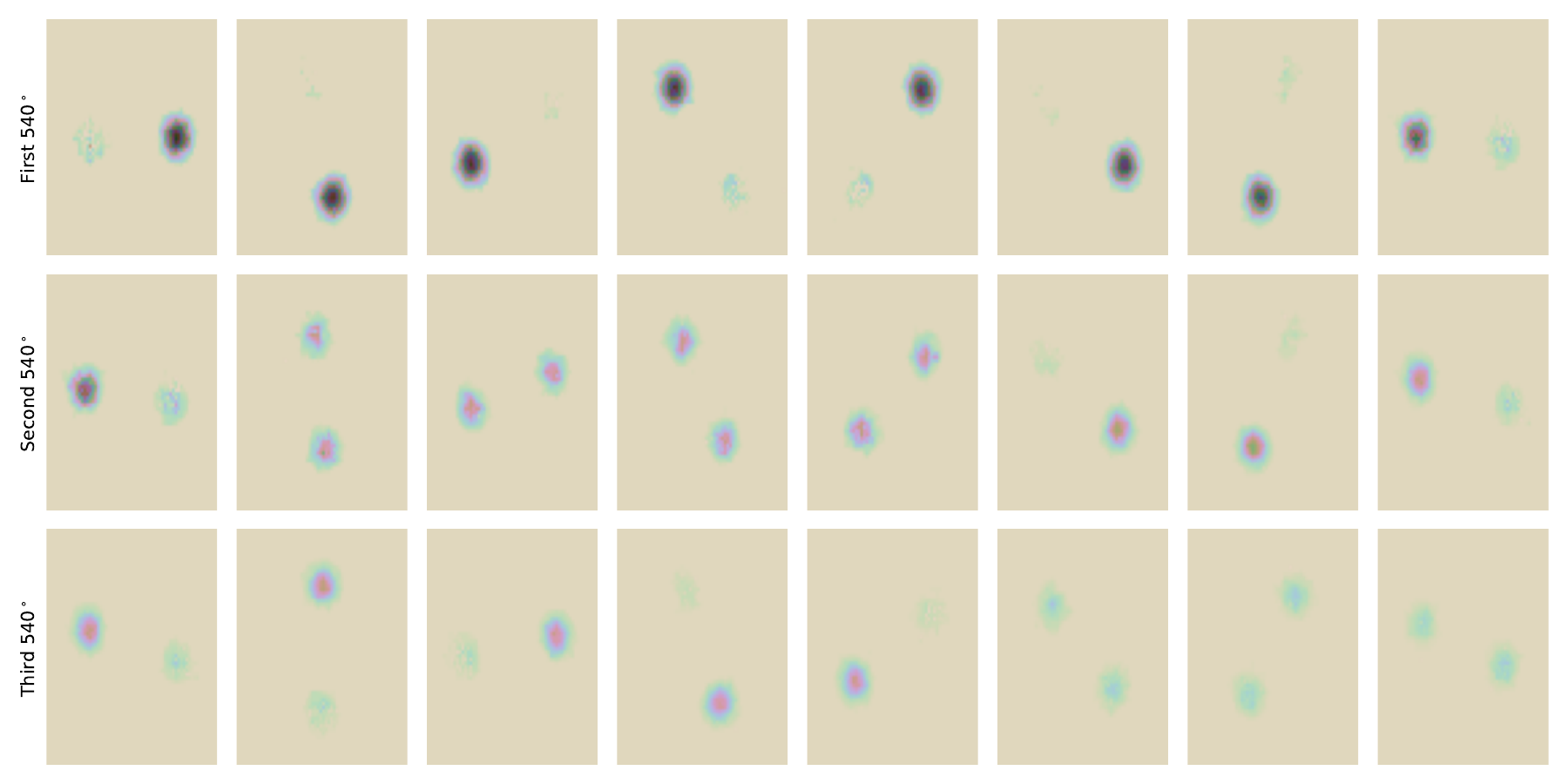}
    \caption{Extrapolation of the learned dynamics using AEDy (no SVD) on the Rotation of a 2D Gaussian dataset to $3$ times the training time horizon (first line is training).}
    \label{fig:rotated_gaussian_extrapolation_noRR}
\end{figure}

While the ability of RRAEDy to perform accurate predictions has been demonstrated in the main paper, we further investigate the model's extrapolation capabilities beyond the training time horizon. To do so, we extrapolate the learned dynamics to $5$ times the training time horizon for the Van der Pol oscillator and $3$ times for the rotated gaussian dataset. The results can be seen in Figures \ref{fig:vdp_extrapolation} and \ref{fig:rotated_gaussian_extrapolation_RR}. For comparison, we also include the extrapolation results using a standard AEDy (without the SVD and rank reduction) on the rotated gaussian dataset in Figure \ref{fig:rotated_gaussian_extrapolation_noRR}.

The figure shows that RRAEDy are capable of accurately extrapolating the learned dynamics well beyond the training time horizon. The model maintains a low error rate even as the predictions extend further into the future, demonstrating its robustness and generalization capabilities. The results further show that the regularization provided by the SVD is necessary for problems with local behavior, as the AEDy without SVD fails to extrapolate accurately in this case. We do not perform similar extrapolation experiments on the other datasets since their dynamics are not periodic, making it difficult to assess the quality of the extrapolated predictions.

\end{document}